%% file: arxiv_version.tex
\documentclass[10pt,twocolumn,letterpaper]{article}
\usepackage[table,xcdraw,dvipsnames]{xcolor}
\usepackage{cvpr}
\usepackage{times}
\usepackage{epsfig}
\usepackage{amsmath}
\usepackage{amssymb}
\usepackage{mathtools}
\usepackage{pifont} 
\usepackage{graphicx, subfigure}
\usepackage{multirow}
\usepackage{color, colortbl}
\usepackage{floatrow}
\usepackage{wrapfig,booktabs}
\usepackage{transparent}
\usepackage{multicol}
\newfloatcommand{capbtabbox}{table}[][\FBwidth]
\usepackage[numbers,sort,compress]{natbib}

\usepackage[backref=page,colorlinks=true,linkcolor=OrangeRed, citecolor=green]{hyperref}

\usepackage[toc,page,header]{appendix}

\usepackage{multirow}
\usepackage{pifont} 
\usepackage{chngcntr}
\cvprfinalcopy 
\input{commands}

\def\cvprPaperID{****} 

\begin{document}

\title{Adaptive Distillation: Aggregating Knowledge from Multiple Paths for Efficient Distillation }

\author{Sumanth Chennupati~~~Mohammad Mahdi Kamani~~~Zhongwei Cheng~~~Lin Chen\\
WYZE Labs AI Team, Kirkland, WA, USA \\
Email: \{schennupati, mkamani, zchen, lchen\}@wyze.com
}

\maketitle

\input{0-abstract}
\input{1-introduction}
\input{2-related}
\input{3-baselines}

\input{5-experiment}
\input{6-conclusion}

\small{
\bibliographystyle{abbrvnat}
\bibliography{egbib}
}

\newpage
\appendix
\input{7-appendix}

\end{document}

%% file: commands.tex
\usepackage{epsf}
\usepackage{graphics}
\usepackage{psfrag}

\usepackage{color}

\usepackage{mathtools}
\usepackage{amsfonts}
\usepackage{amsthm}
\usepackage{amsmath}
\usepackage{amssymb}

\long\def\comment#1{}

\newcommand{\bm}[1]{\mathbf{#1}}

\definecolor{mygreen}{rgb}{0.31, 0.78, 0.47}
\definecolor{myred}{rgb}{0.9, 0.17, 0.31}
\newlength\MAX  
\newcommand*\Chartg[1]{\rlap{\textcolor{gray!30}{\rule{\MAX}{2ex}}}\transparent{0.6}{\textcolor{mygreen}{\rule{#1\MAX}{2ex}}}}

\newcommand*\Chartr[1]{\rlap{\textcolor{gray!30}{\rule{\MAX}{2ex}}}\transparent{0.6}{\textcolor{red}{\rule{#1\MAX}{2ex}}}}
\newcommand*\Chartrr[1]{\rlap{\textcolor{gray!30}{\rule{\MAX}{2ex}}}\transparent{0.6}{\textcolor{myred}{\rule{#1\MAX}{2ex}}}}

\newcommand*\Chartb[1]{\rlap{\textcolor{gray!30}{\rule{\MAX}{2ex}}}\transparent{0.6}{\textcolor{cyan}{\rule{#1\MAX}{2ex}}}}

%% file: 0-abstract.tex
\begin{abstract}
Knowledge Distillation is becoming one of the primary trends among neural network compression algorithms to improve the generalization performance of a smaller student model with guidance from a larger teacher model. This momentous rise in applications of knowledge distillation is accompanied by the introduction of numerous algorithms for distilling the knowledge such as soft targets and hint layers. Despite this advancement in different techniques for distilling the knowledge, the aggregation of different paths for distillation has not been studied comprehensively. This is of particular significance, not only because different paths have different importance, but also due to the fact that some paths might have negative effects on the generalization performance of the student model. Hence, we need to adaptively adjust the importance of each path to maximize the impact of distillation on the student model. In this paper, we explore different approaches for aggregating these different paths and introduce our proposed adaptive approach based on multitask learning methods. We empirically demonstrate the effectiveness of the proposed approach over other baselines on the applications of knowledge distillation in classification, semantic segmentation, and object detection tasks.
\end{abstract}

%% file: 1-introduction.tex
\section{Introduction}
\label{sec:intro}
The promising advancements of deep learning models in various AI tasks are dominantly depending on their large and complex model structures, which grants them boosted generalization capabilities on test data. However, the benefits of these achievements are limited in resource-constrained systems such as mobile devices, low power robots, etc. Different model compression algorithms~\cite{song2015learning,huang2018learning,blalock2020state,song2016deep,Jaderberg_2014,kim2016compression,khan2020learning} have been proposed to reduce the complexity of such larger models for these systems. Among different compression algorithms, distilling the knowledge from a larger model to a smaller one has shown to be highly effective and advantageous~\cite{hinton2015distilling,lopez2015unifying,polino2018model}. Beyond compression capabilities, knowledge distillation has also been a primary motive for different techniques that transfer the knowledge between models such as domain adaptation~\cite{long2016unsupervised,kundu2019adapt}.

With the upsurge in the applications of knowledge distillation in different domains, various methods were introduced for distilling the knowledge from a teacher to a student. These approaches are applied to different parts (\emph{i.e.} distillation paths) of the models such as output logits~\cite{hinton2015distilling} and hidden layers' feature maps~\cite{romero2014fitnets}. Although most of these distillation paths could improve the generalization performance of the student model, a naive combination of these paths might have negative effects on the performance of the student model. For instance, when we use knowledge distillation for an object detector, we can distill the knowledge from their backbone feature maps~\cite{zhang2021improve} or its bounding box generator~\cite{zheng2021LD}. As it can be inferred from Table~\ref{tab:motive}, naively combining backbone features' distillation paths with those from the bounding box generator can degrade the performance when compared to the bounding box generator alone. This phenomenon indicates the necessity for a systematic approach towards the aggregation of different paths to gain the most from the knowledge distillation process.

\begin{table}[!tb]

	\resizebox{.8\textwidth}{!}{%
		\begin{tabular}{lcccc}
			\toprule
			                  & Feats        & Box         & mAP           & mAP\textsubscript{50} 
			\\
			\midrule
			Student           &              &              & 38.1          & 59.9                  
			\\\midrule
			Single        & $\checkmark$ &              & 39.8          & 62.3                  \\
			Single       &              & $\checkmark$ & 39.6          & 62.4                  \\
			Hand-tuned        & $\checkmark$ & $\checkmark$ & 39.7          & 61.4                  \\
			\textbf{Adaptive (ours)} & $\checkmark$ & $\checkmark$ & \textbf{40.2} & \textbf{62.9}         \\ \midrule
			Teacher           &              &              & 42.9          & 65.7                  \\
			\bottomrule
		\end{tabular}
		\caption{Comparing a hand-tuned versus adaptive approach to aggregate distillation paths for an object detection task on Cityscapes \cite{Cordts2016Cityscapes}. For more details refer to Section~\ref{sec:exp}.}\label{tab:motive}
	}

\end{table}

In this paper, we propose an adaptive approach to learn the importance of each path during the distillation and training process. This approach is inspired by multitask learning methods~\cite{caruana1997multitask}, where we consider each path as a separate task that we want to optimize the model for. Using our adaptive approach, we can mitigate the negative effects mentioned above and benefit the most while aggregating from multiple paths of distillation. As it can be seen in Table~\ref{tab:motive}, our proposed Adaptive approach can aggregate these two paths and surpasses both of them in improving the performance of the student model. In addition to our adaptive approach, we propose another baseline method based on multiobjective optimization, where it reduces the aggregation problem to a multi-criteria optimization and intends to find its Pareto stationary solutions. This approach seems to be more effective than naive aggregation methods, but cannot outperform our proposed adaptive distillation.

The main contributions of this paper can be summarized as follows:
\begin{itemize}
	\item We propose an adaptive distillation approach, inspired by multitask learning methods for efficient aggregation of different distillation paths. We provide a general optimization formulation for this problem and reduce different methods using this formulation.
	\item We introduce multiobjective optimization for this problem as a baseline approach, which can be more effective than naive aggregation approaches.
	\item We conduct extensive comparison between our approach and other baseline methods in different tasks of image classification, semantic segmentation, and object detection.
\end{itemize}

%% file: 2-related.tex
\section{Related Work}\label{sec:related}
Knowledge distillation has become one of the prevailing approaches in model compression to improve the generalization performance of a smaller student model using the knowledge from a richer teacher model. In this section, we provide an overview of proposals in this domain, as well as methods we utilize to aggregate different paths of knowledge distillation.

\textbf{Knowledge Distillation}
The concept of knowledge distillation for neural networks was first introduced by~\citet{hinton2015distilling} to distill the knowledge from a teacher to a student model by minimizing the distance between their soft targets. This idea has been expanded to the hidden layers' feature maps by the introduction of Hint layers by~\citet{romero2014fitnets}. The idea of knowledge distillation and its variants has extensively employed in various problems such as compression~\cite{hinton2015distilling,mishra2017apprentice,ashok2017n2n,wang2019distilling,chen2017learning,polino2018model}, knowledge transfer~\cite{zeng2019wsod2,radosavovic2018data,wang2018dataset}, and federated learning~\cite{seo2020federated,lin2020ensemble,haddadpour2021federated}. It has also been explored in different tasks such as image classification~\cite{hinton2015distilling,mirzadeh2020improved,yuan2020revisiting}, object detection~\cite{wang2019distilling,chen2017learning,zhang2021improve,dai2021general}, semantic segmentation~\cite{he2019knowledge,michieli2021knowledge,xie2018improving}, and graph neural networks~\cite{lassance2020deep,ma2019graph} to name but a few. More detailed discussions regarding knowledge distillation approaches and state-of-the-art methods in this domain can be found in~\cite{gou2021knowledge,wang2021knowledge}.

\textbf{Ensemble of Teachers} The problem of aggregating different knowledge paths from multiple teachers has been the primary topic of several studies. Most studies investigate how to effectively combine the ensemble of teachers' outputs (logits) for distillation to a student model using weighted averaging~\cite{allen2020towards,fukuda2017efficient,lan2018knowledge}. Different approaches have been proposed for finding these weights such as the teachers' confidence score~\cite{xiang2020learning} or the similarity between two models' inputs~\cite{zhang2018better}. \citet{zhang2018deep} introduce the reverse problem, where they can improve the teacher with an ensemble of student models using symmetrical KL divergence. \citet{piao2020a2dele} introduce A2dele to combine depth and RGB networks using a confidence-based weights for distillation paths.
In some other studies, the problem of an ensemble of teachers on hint layers' feature maps has been studied~\cite{park2019feed,liu2019knowledge}. However, the number of these proposals is limited due to the challenging nature of this problem because of the non-aligning size of feature maps in different teachers. As it is mentioned by~\cite{wang2021knowledge}, finding a systematic approach to calculate the degree of efficiency for each distillation path in this problem remains open and challenging. To the best of our knowledge this is one of the first attempts to adaptively combine distillation paths with different nature in a training procedure.

\textbf{Multiobjective Optimization} An efficient way to aggregate different distillation paths is to consider the problem as multiobjective optimization, and hence, benefit from approaches proposed in this domain. There are various methods to solve a multiobjective optimization, however, for the sake of efficiency of the approach, we will use a first-order gradient-based method like the ones proposed in~\cite{desideri2012multiple,sener2018multi,mahapatra2020multi}. Using these approaches, we can converge to the Pareto stationary of the problem, where no other solution can dominate that solution.

\textbf{Multitask Learning} Another way of aggregating knowledge from different distillation paths is to adjust the importance of each path by scaling their losses. Early works in multitask learning (MTL)~\cite{caruana1997multitask, teichmann2018multinet, Kokkinos2017ubernet, Chennupati_2019} use a weighted arithmetic sum of individual task losses. These weights are hand-tuned or manually searched and remain static throughout the training process. Later, \citet{liu2019end} and \citet{guo2018dynamic} propose to adjust these weights by inspecting the change in loss or difficulty of each task as the training progresses. GradNorm~\cite{Chen2018GradNormGN} proposes to normalize gradients from different losses to a common scale during backpropagation. \citet{kendall2017multi} and \citet{leang2020dynamic} propose to consider different task weights as parameters and learn them using backpropagation. 

%% file: 3-baselines.tex
\section{Methods}\label{sec:base}
Distilling knowledge from a teacher to a student can take place at any stage of a model from initial feature maps to final logits of the output. Not all of them have the same effect on boosting the generalization performance of the student model. In fact, in practice, it can be seen that some of these distillation paths might hurt the generalization of the student models. Hence, it is of paramount importance to take these effects into consideration when updating the parameters based on each of these paths. In this section, we first introduce the problem formulation. Next, using this formulation, we provide few baseline approaches on how to aggregate different paths of distillation during the training procedure.

\subsection{Problem Formulation} \label{sec:pf}
In order to better formulate the multiple paths distillation problem at hand, we first describe the main learning task. Universally, in many forms of supervised learning tasks, the primary objective is to find an optimal prediction mapping, given a training dataset $\mathcal{T}$, with $N$ training samples. The mapping is between the input feature space $\mathcal{X}$ and the target space $\mathcal{Y}$, whether it is a classification, semantic segmentation, or object detection task. In this case, each sample is presented with the intersection of these two spaces denoted by $(\bm{x}^{(i)},y^{(i)}) \in \mathcal{X}\times\mathcal{Y}, i\in[N]$. Deep neural networks aim at representing this mapping using an $M$-layer neural network model, where each layer $l$ is represented by a parameter set of $\bm{w}_l \in \mathbb{R}^{d_l}$ and applies the transformation of $f_l\left(.;\bm{w}_l\right)$ on its input. The set of all parameters of the network is denoted by $\bm{w} = \left\{\bm{w}_1,\ldots,\bm{w}_M\right\} \in \mathbb{R}^d$, where $d=\sum_{i\in[M]}d_i$. Thus the main objective of this supervised task is to minimize the empirical risk on the training data, which is:
\vspace{-0.25cm}
\begin{equation}
	\mathcal{L}\left(\mathcal{T};\bm{w}\right) = \frac{1}{N} \sum_{i\in[N]} \ell\left(\bm{x}_i,y_i;\bm{w}\right),
\vspace{-0.25cm}
\end{equation}
where $\ell\left(.,.;.\right)$ is the loss on each sample such as cross entropy loss. In knowledge distillation frameworks the ultimate goal is for the student to imitate the teacher's output features in different layers. 
For instance, in the primary form of the knowledge distillation using soft targets~\cite{hinton2015distilling}, these outputs are the soft logits of the two models. Whereas in hint layers~\cite{romero2014fitnets} and attention transfer \cite{zagoruyko2016paying}, these features are the middle layers' outputs. 
It should be noted that each distillation path only affects a subset of the parameters in the student model (unless for the soft target or equivalent, where all the parameters of the student model are affected). Hence, the general form of knowledge distillation loss for each path between the $j$-th layer on the student model and the $k$-th layer on the teacher model can be formulated as:
\vspace{-0.25cm}
\begin{equation}\label{eq:kd_loss}
	\mathcal{L}_{KD}^{jk}\left(\mathcal{X};\bm{w}^\text{S},\bm{w}^\text{T}\right) = \frac{1}{N} \sum_{i\in[N]} \ell_{KD}^{jk} \left(\bm{x}_i;\bm{w}^\text{S}_{\leq j},\bm{w}^\text{T}_{\leq k}\right),
\vspace{-0.25cm}
\end{equation}
where $\bm{w}^\text{S}$ and $\bm{w}^\text{T}$ are the student and teacher model parameters, respectively. The set of parameters for the layers up to layer $j$ is denoted by $\bm{w}_{\leq j}$. The loss for each sample is calculated based on the loss function $\ell_{KD}^{jk}\left(.;.,.\right)$. For instance, if the path is the soft target, the loss is the KL-divergence between two soft logits, and if it is a hint layer, the loss could be a simple euclidean distance~\cite{romero2014fitnets}. For hint layers an \textit{adaptation} layer $\bm{w}_A$ might be necessary to match the spatial or channel size of the feature maps on the student model to that of the teacher model. The parameters of this layer will be tuned using the distillation loss defined in Eq.~(\ref{eq:kd_loss}). Thus, if we consider $K$ paths for distillation with their own defined empirical loss function as in Eq.~(\ref{eq:kd_loss}), we can create a distillation loss vector denoted by $\mathrm{\bm{f}}_{KD} \left(\mathcal{X};\bm{w}^\text{S},\bm{w}^\text{T}\right) \in \mathbb{R}^K$. Then, the overall optimization can be written as:
\vspace{-0.25cm}
\begin{equation}\label{eq:comp_loss}
	\min_{\bm{w}^\text{S} \in \mathbb{R}^{d_s}} \mathcal{L}\left(\mathcal{T};\bm{w}^\text{S}\right) + \alpha \cdot \bm{v}^\top \mathrm{\bm{f}}_{KD} \left(\mathcal{X};\bm{w}^\text{S},\bm{w}^\text{T}\right)
\vspace{-0.25cm}
\end{equation}
where $\alpha \in \left[0,1\right]$ is the weight indicating the importance of the distillation loss compared to the main empirical loss. $\bm{v}\in \mathbb{R}^K$ is the vector indicating the weight for each distillation path. For the current setting, we only consider the linear combination of losses between these paths; however, the nonlinear combination can be investigated in future studies. Next, we will describe how we can combine these losses during the training to distill the knowledge from the teacher model to the student model.

    \begin{figure*}[!tb]
        \centering
        \subfigure[Single Distillation] {\label{fig:kd_loss}\includegraphics[width=0.235\textwidth]{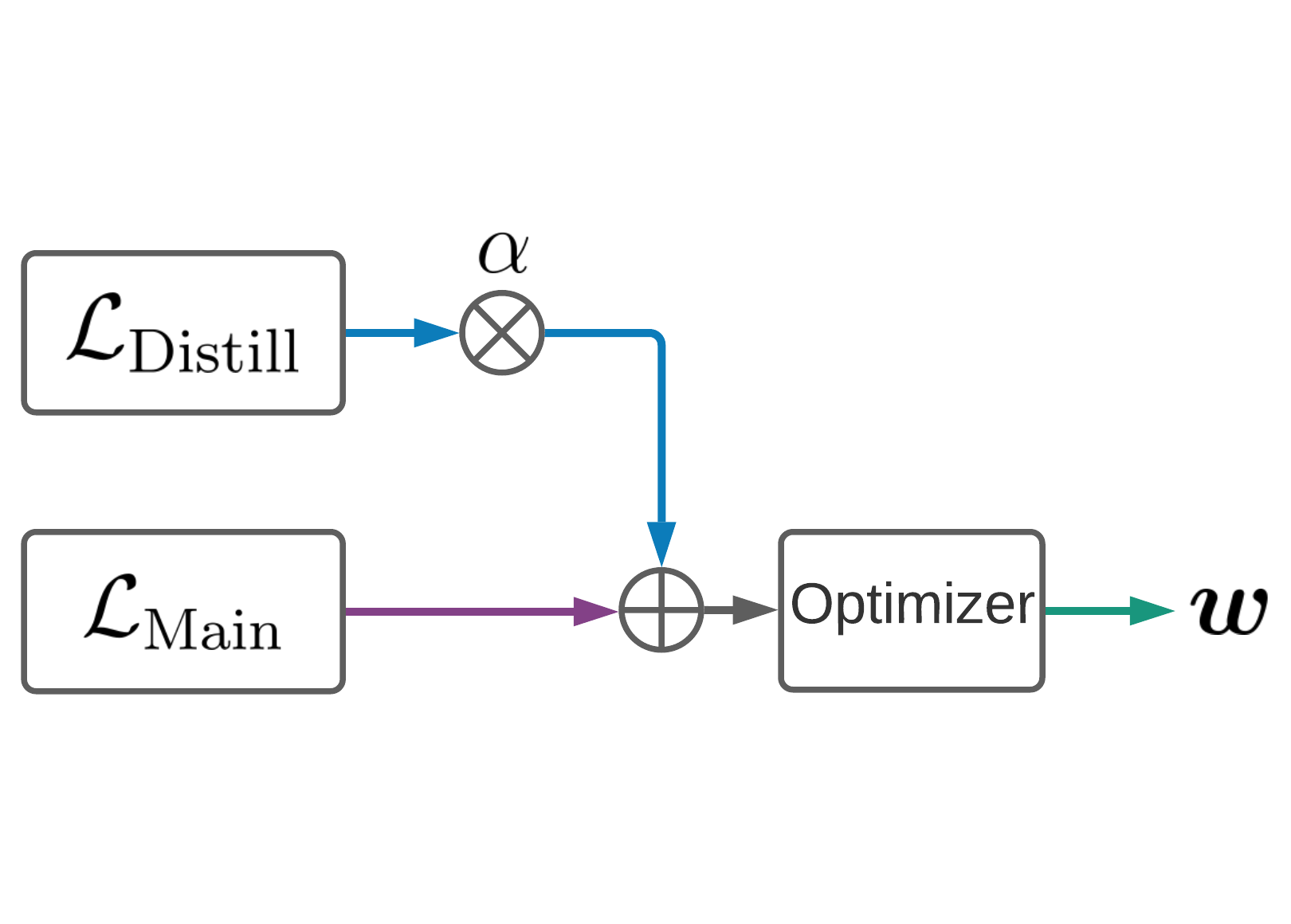}} \hspace{1mm}
        \subfigure[Hyperparameter Optimization] {\label{fig:mtl}\includegraphics[width=0.235\textwidth]{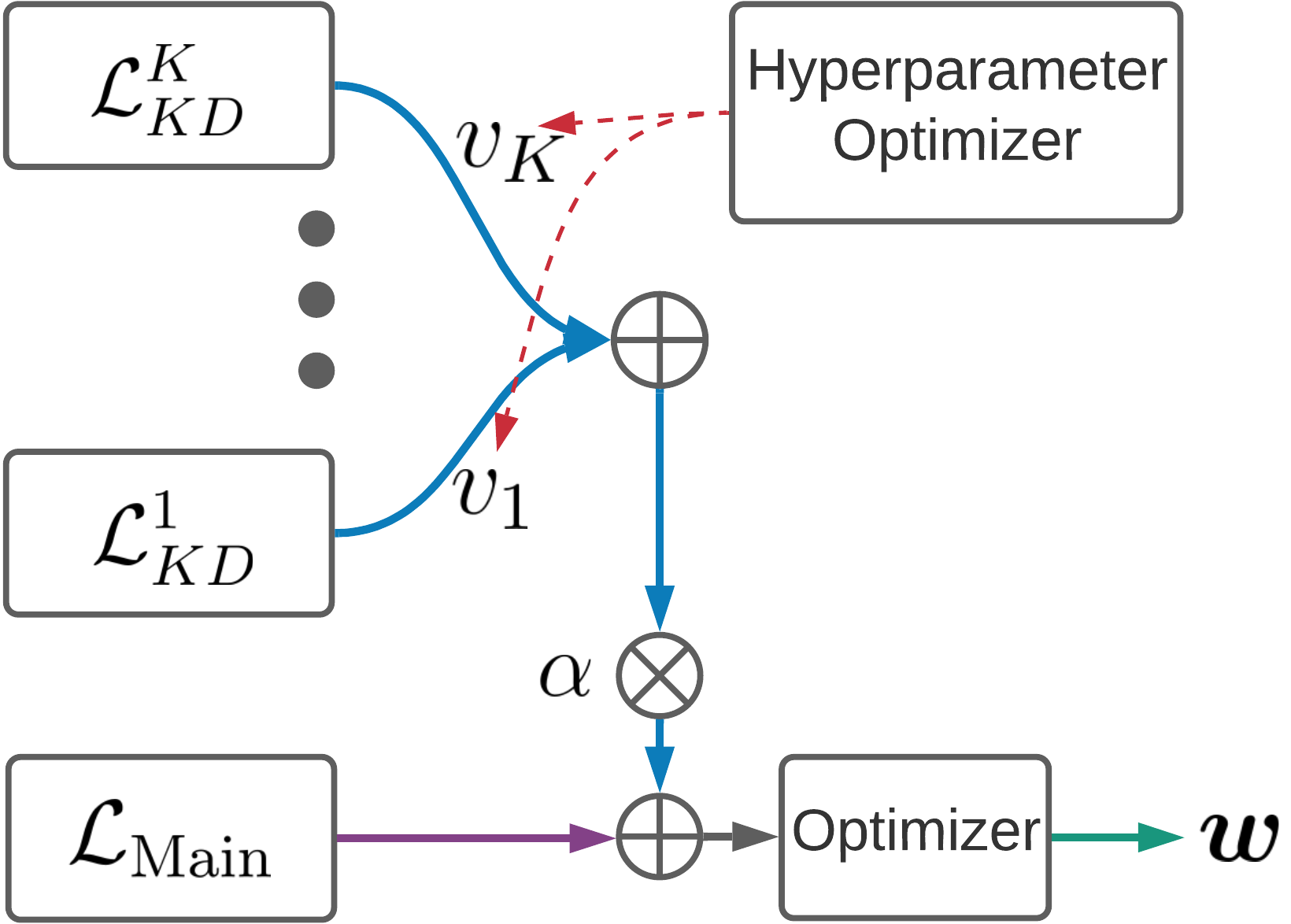}}\hspace{1mm}
        \subfigure[Multiobjective Optimization]{\label{fig:moo}\includegraphics[width=0.235\textwidth]{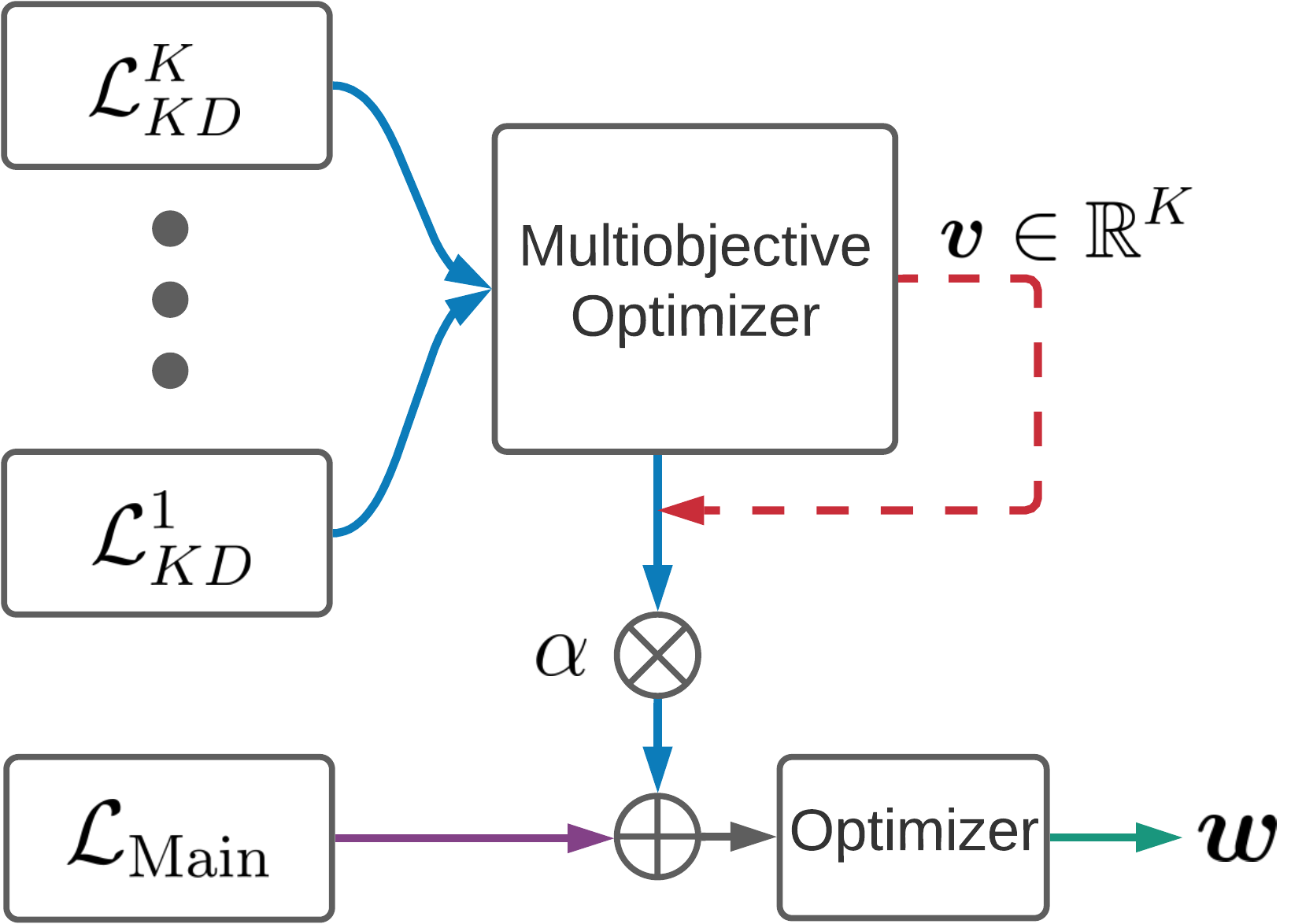}} \hspace{1mm}
        \subfigure[Adaptive Distillation.]{\label{fig:adaptive}\includegraphics[width=0.235\textwidth]{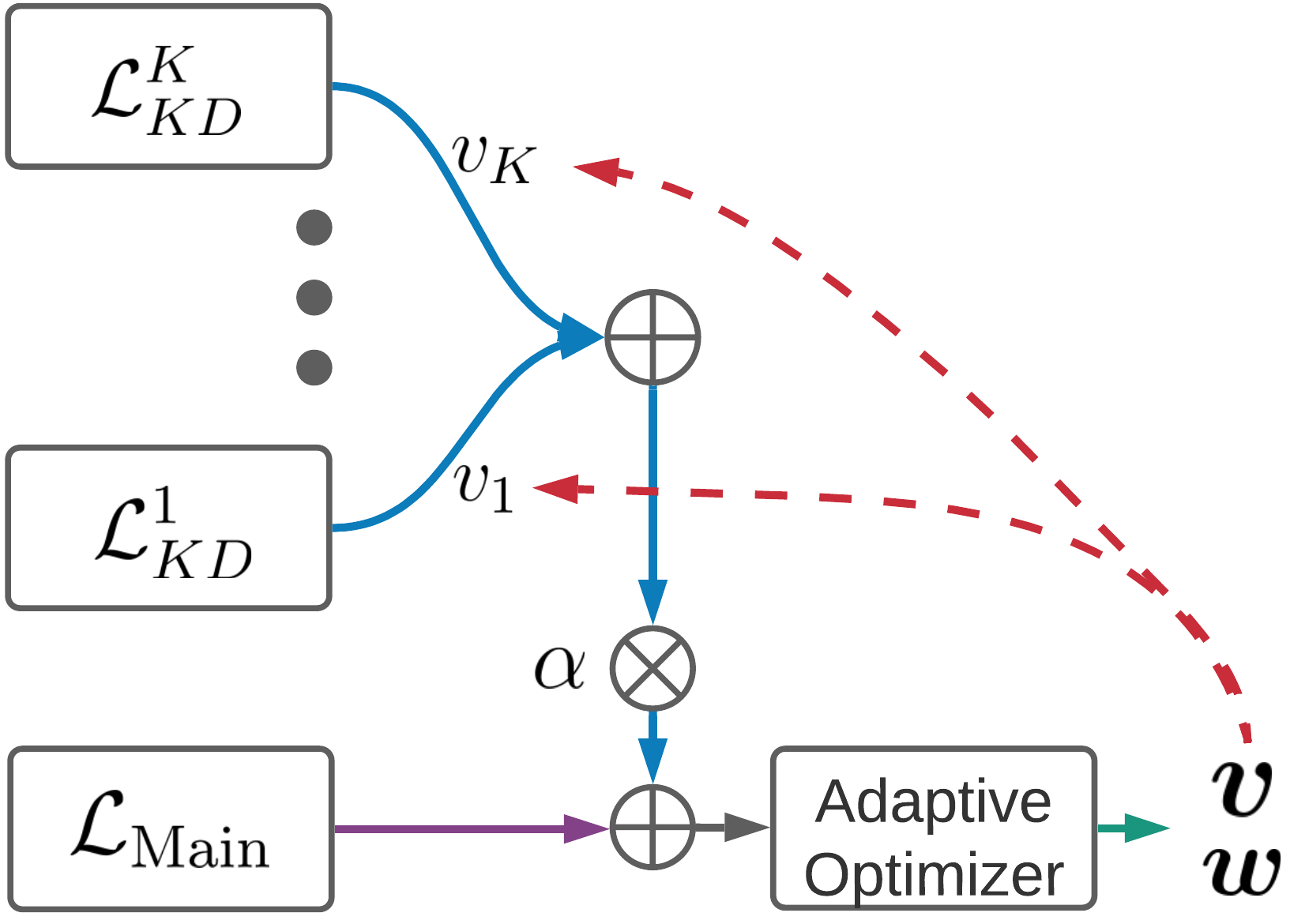}}
        \caption{Schematic of different aggregation approaches for knowledge distillation. (a) The single distillation approach. (b) Hyperparameter optimization approach to find the optimal weight for each path before the training. (c) Multiobjective optimization to find the descent direction for multiple paths at every iteration. (d) The proposed adaptive distillation to adaptively learn the weights using the main optimizer of the problem.}
        \label{fig:Loss_functions}
        \vspace{-0.35cm}
    \end{figure*}

\subsection{Equal Weights} The most naive and common form of combining these distillation paths is to consider all of them equally weighted during the optimization. Considering the overall objective of this problem in Eq.~(\ref{eq:comp_loss}), for this case, we should consider all distillation paths' weights equal, which is $\bm{v} = \left[1,\ldots1\right]^\top$. Although this might work in some cases, due to different scales of these losses, in most cases we need to adjust the weights accordingly.

\subsection{Hyperparameter Optimization} Another baseline for this problem is to consider the weights for each distillation path as a new hyperparameter that we need to optimize before the training procedure as shown in Figure~\ref{fig:mtl}. In this case, the weight vector $\bm{v}$ is considered to be a hyperparameter that needs to be tuned. This can be done using different hyperparameter optimization approaches such as grid search, random search, or bilevel optimization~\cite{franceschi2018bilevel}.

\subsection{Multiobjective Optimization} If we look closely at the objective of the optimization in Eq.~(\ref{eq:comp_loss}), it can be evidently reformulated into a multiobjective optimization as shown in Figure~\ref{fig:moo}. In multiobjective optimization, the goal is to find a Pareto stationary point, where no other local solution can dominate that solution in any of the objectives in the task. Most of the first-order gradient-based approaches, based on this notion, aim at finding the direction at every step that is not harming any of the objectives until such direction cannot be found. In this way, all the objectives are treated equally and we try to find a Pareto stationary point of the problem.

The main challenge of using multiobjective optimization in our task is that the fundamental goal of the main empirical risk and distillation losses are not the same in the essence, and the former is more important than the latter. A solution is to use preference-based approaches~\cite{mahapatra2020multi, kamani2021pareto}, to put more emphasis on the main objective of the learning. Another way is to consider the optimization problem of multiple paths distillation as a multiobjective optimization to find the best weights $\bm{v}$ and then combine it with the main empirical risk with the weight of $\alpha$. In this way, at every iteration, we first find a descent direction for all distillation losses, and then combine it with the gradients of the main learning objective using the weight $\alpha$. To find the descent direction for distillation losses at every iteration we use approaches introduced in different studies~\cite{miettinen2012nonlinear, desideri2012multiple,sener2018multi,lin2019pareto,mahapatra2020multi, kamani2019efficient,kamani2020multiobjective}. 

To do so, for each distillation path, we compute the gradients of the parameters of the student model affected by that loss on the mini-batch $\bm{\xi}$. Then for each path $i$ we combine the gradients of the parameters from different layers together in a vector $\mathrm{\bm{g}}_i\left(\bm{\xi};\bm{w}^\text{S}_{\leq j_i},\bm{w}^\text{T}_{\leq k_i}\right)$, where $j_i$ and $k_i$ are the layer indices of student and teacher models, respectively, for the $i$-th distillation path. Since in each distillation path not all of the parameters in the student model are involved, we will use zero gradients for the parameters not involved in a path when gathering all the gradients. Then, by solving the following quadratic optimization, we will find the optimal weights of $\bm{v}$ for the current mini-batch as:
\vspace{-0.25cm}
\begin{equation}\label{eq:quad}
	\bm{v}^* \in \arg\underset{\bm{v} \in \Delta_K}{\min} \left\|\sum_{i \in [K]} v_i\mathrm{\bm{g}}_i\left(\bm{\xi};\bm{w}^\text{S}_{\leq j_i},\bm{w}^\text{T}_{\leq k_i}\right)\right\|_2^2,
\end{equation}
where $\Delta_K = \left\{p_i | 0 \leq p_i \leq 1, \sum_{i\in[K]}p_i = 1\right\}$ is a $K$-dimensional simplex. It has been shown that using the $\bm{v}^*$ from Eq.~(\ref{eq:quad}) the resulting direction is descent for all distillation paths~\cite{sener2018multi}.

\subsection{Adaptive Distillation}\label{sec:method}
Inspired by multitask learning~\cite{kendall2017multi, leang2020dynamic}, we intend to learn the importance of each distillation path adaptively by considering each path as a separate task in the learning. To do so, in addition to the parameters in Eq.~(\ref{eq:comp_loss}), we introduce a new set of proxy parameters $\bm{z} = [z_1, \ldots, z_K]$ to estimate
$\bm{v} = [e^{-z_1}, \ldots, e^{-z_K}]$ as shown in Figure \ref{fig:adaptive}. Thus, we update the objective of the optimization in Eq.~(\ref{eq:comp_loss}) as:
\begin{align}\label{eq:adaptive_loss}
	\min_{\bm{w}^\text{S} \in \mathbb{R}^{d_s}, \bm{z}\in\mathbb{R}^K} &\mathcal{L}\left(\mathcal{T};\bm{w}^\text{S}\right) \\\nonumber
	&+ \underbrace{\alpha \cdot (\bm{v}^\top \mathrm{\bm{f}}_{KD} \left(\mathcal{X};\bm{w}^\text{S},\bm{w}^\text{T}\right) + \textstyle \sum_{i\in[K]}z_i}_{f_\text{d}\left(\mathcal{X};\bm{w}^\text{S},\bm{w}^\text{T},\bm{z}\right)},)
\end{align}
where the last two terms define the distillation loss in terms of the model parameters and the proxy parameters $\bm{z}$. Expressing $v_i$ as $e^{-z_i}$ ensures $v_i > 0 \;\;\forall z_i \in \mathbb{R}$. The term $\sum_{i\in[K]} z_i$ acts as a regularization to prevent $z_i$ to converge to larger values that decreases $v_i$, and thereby, vanishes the second loss term. To understand how $z_i$ gets updated, we inspect the gradients of the distillation loss in Eq.~(\ref{eq:adaptive_loss}) with respect to $z_i$ on mini-batch of $\xi$ denoted by:
\begin{equation}\label{eq:grad}
	\left.\frac{\partial f_\text{d}\left(\mathcal{X};\bm{w}^\text{S},\bm{w}^\text{T},\bm{z}\right)}{\partial z_i}\right|_{\xi} = \alpha\left[1 - e^{-z_i} \cdot \ell_{KD}^{j_ik_i} \left(\bm{\xi};\bm{w}^\text{S}_{\leq j_i},\bm{w}^\text{T}_{\leq k_i}\right)\right]
\end{equation}
Hence, if we have access to the true loss (full batch), based on the first-order optimality condition resulting from Eq.~(\ref{eq:grad}), we can infer that the optimal value for $v_i$ would be equal to the inverse of its corresponding loss. This means that scaling the gradients of each loss at every iteration with the inverse of their true loss will give us the optimal results. However, due to the infeasibility of the true loss at every iteration, and since we are using stochastic gradient descent for the optimization, we will use the stochastic gradient in Eq.~(\ref{eq:grad}) to update $z_i$ values at every iteration. We expect this will converge to the optimal values for each path.

%% file: 5-experiment.tex
\begin{table*}[!htb]
    \addtolength{\tabcolsep}{-4pt}
    \centering
    \resizebox{\textwidth}{!}{
        \begin{tabular}{lccllllllll}

            \toprule
                              & \multicolumn{2}{c}{\textbf{Path(s)}} & \multicolumn{2}{c}{\textbf{CIFAR-10}} & \multicolumn{2}{c}{\textbf{CIFAR-100}} & \multicolumn{2}{c}{\textbf{ImageNet-200}} & \multicolumn{2}{c}{\textbf{ImageNet-1K}}                                                                                                                                              \\
            \cmidrule(r){2-3} \cmidrule(r){4-5} \cmidrule(r){6-7} \cmidrule(r){8-9}\cmidrule(r){10-11}

                              & \textbf{AT}                          & \textbf{ST}                           & \textbf{top1 err} ($\downarrow$)       & \textbf{top1 agr} ($\downarrow$)          & \textbf{top1 err} ($\downarrow$) & \textbf{top1 agr} ($\downarrow$) & \textbf{top1 err} ($\downarrow$) & \textbf{top1 agr} ($\downarrow$) & \textbf{top1 err} ($\downarrow$) & \textbf{top1 agr} ($\downarrow$)\\ \midrule
            Student           &                                      &                                       & 5.19 \Chartr{1}                        & \multicolumn{1}{c}{-}                     & 24.84 \Chartr{1}                 & \multicolumn{1}{c}{-}            & 42.34 \Chartr{1}                 & \multicolumn{1}{c}{-}         & 30.10 \Chartr{1} & \multicolumn{1}{c}{-}    \\ \midrule
            Single            & $\checkmark$                         &                                       & 4.51 \Chartr{0.24}                     & 3.20 \Chartb{0.15}                        & 22.87 \Chartr{0.6}               & 19.04 \Chartb{0.83}              & 39.80 \Chartr{0.47}              & 35.07 \Chartb{0.95}      & 29.42 \Chartr{0.89} &   21.80  \Chartb{1.0}     \\
            Single            &                                      & $\checkmark$                          & 5.10 \Chartr{0.9}                      & 4.06 \Chartb{0.86}                        & 22.67 \Chartr{0.56}              & 19.71 \Chartb{1}                 & 40.24 \Chartr{0.56}              & 35.38 \Chartb{1.00}    & 29.09 \Chartr{0.89}&  21.66    \Chartb{0.9}    \\
            Hand-tuned        & $\checkmark$                         & $\checkmark$                          & 4.75 \Chartr{0.51}                     & 3.74 \Chartb{0.6}                         & 21.47 \Chartr{0.31}              & 17.40 \Chartb{0.41}              & 39.00 \Chartr{0.56}              & 32.89 \Chartb{0.59}          & \multicolumn{1}{c}{-}& \multicolumn{1}{c}{-}                 \\
            Equal             & $\checkmark$                         & $\checkmark$                          & 5.05 \Chartr{0.84}                     & 4.23 \Chartb{1}                           & 21.75 \Chartr{0.37}              & 18.53 \Chartb{0.7}               & 38.25 \Chartr{0.14}              & 29.83 \Chartb{0.08}         & 28.73  \Chartr{0.79} &   20.52 \Chartb{0.15}\\
            Multiobjective    & $\checkmark$                         & $\checkmark$                          & 4.65 \Chartr{0.4}                      & 3.63 \Chartb{0.5}                         & 21.58 \Chartr{0.33}              & 18.09 \Chartb{0.59}              & 39.75 \Chartr{0.46}              & 33.22 \Chartb{0.64} & \multicolumn{1}{c}{-}& \multicolumn{1}{c}{-}             \\
            \textbf{Adaptive} & $\checkmark$                         & $\checkmark$                          & \textbf{4.39} \Chartr{.0}              & \textbf{3.12} \Chartb{0.00}               & \textbf{20.04} \Chartr{0}        & \textbf{15.89} \Chartb{0.0}      & \textbf{37.68} \Chartr{0.0}      & \textbf{29.46} \Chartb{0.0}  & \textbf{28.39} \Chartr{0.74} &   \textbf{20.29} \Chartb{0.0}\\ \midrule
            Teacher           &                                      &                                       & 4.63 \Chartr{0.38}                     & \multicolumn{1}{c}{-}                     & 20.10 \Chartr{0.03}              & \multicolumn{1}{c}{-}            & 40.81 \Chartr{0.68}              & \multicolumn{1}{c}{-}    & 23.45 \Chartr{0}& \multicolumn{1}{c}{-}        \\ \bottomrule
        \end{tabular}
    }
    \caption{Validation results for different knowledge distillation methods on CIFAR-10, CIFAR-100, ImageNet-200 and ImageNet-1K. We report top1 classification error (\%) for the student model and top1 agreement error (\%) defined in~\cite{stanton2021does} (i.e, classification error between teacher and student). Adaptive distillation achieves the best results, even better than the teacher model, in all datasets. $(\downarrow)$ indicates lower the better. }
    \label{tab:image-classification-results}
    \addtolength{\tabcolsep}{1pt}
    
\end{table*}

\section{Empirical Studies}\label{sec:exp}
In this section, we explore knowledge distillation with multiple paths from a teacher using the approaches mentioned in Section~\ref{sec:base}, in comparison with our proposed adaptive distillation. For most of the experiments, unless specified, we use ResNet50~\cite{he2016deep} as a teacher and ResNet18 as a student. We chose attention transfer (AT)~\cite{zagoruyko2016paying} for feature distillation paths, and soft target (ST)~\cite{hinton2015distilling} for logit distillation, as two paths for distillation. For attention transfer, we use the sum of squared differences between attention maps of features from layers 2 through 5 of ResNet50 and ResNet18. We use $\bm{v}$ = $[v_{AT}, v_{ST}]$ = $[1, 1]$ for `Equal' weights baseline and chose $\bm{v}$ = $[1000, 0.1]$ for `Hand-tuned' weights baseline based on our grid search hyperparameter tuning. Multiobjective, and Adaptive methods initialize weights as $\bm{v}$ = $[v_{AT}, v_{ST}]$ = $[1, 1]$ and aim to learn the best weights during training. We use $\alpha$=1.0 which is the relative importance of aggregated distillation losses compared to the main loss. A more detailed discussion on the experimental setups with additional results are deferred to Appendix~\ref{app:exp}.

\subsection{Results}

\paragraph{Image Classification:} We perform experiments on CIFAR-10  and CIFAR-100~\cite{krizhevsky2009learning}, as well as ImageNet-200~\cite{wu2017tiny} and ImageNet-1K~\cite{ILSVRC15} for the image classification task. We evaluate top1 classification error on the validation dataset along with the top1 agreement error as suggested by~\citet{stanton2021does} for knowledge distillation methods (i.e, classification error between teacher and student). We train our student for 200 epochs with batch size 128 on CIFAR datasets and 100 epochs with batch size 256 on ImageNet-200, ImageNet-1K datasets. We chose an SGD optimizer with an initial learning rate of 0.1 for all datasets. We reduce the learning rate by a factor of 0.1 at [100, 150] epochs for CIFAR-10, [30, 60, 90] for both ImageNet datasets. For CIFAR-100, we reduce the learning rate by 0.2 at [60, 120, 160] epochs. We use random flip augmentation during training in experiments.

\begin{figure*}[!tb]
    \centering
    \subfigure[CIFAR-10]{\label{fig:cifar10_sota}\includegraphics[width=0.32\textwidth]{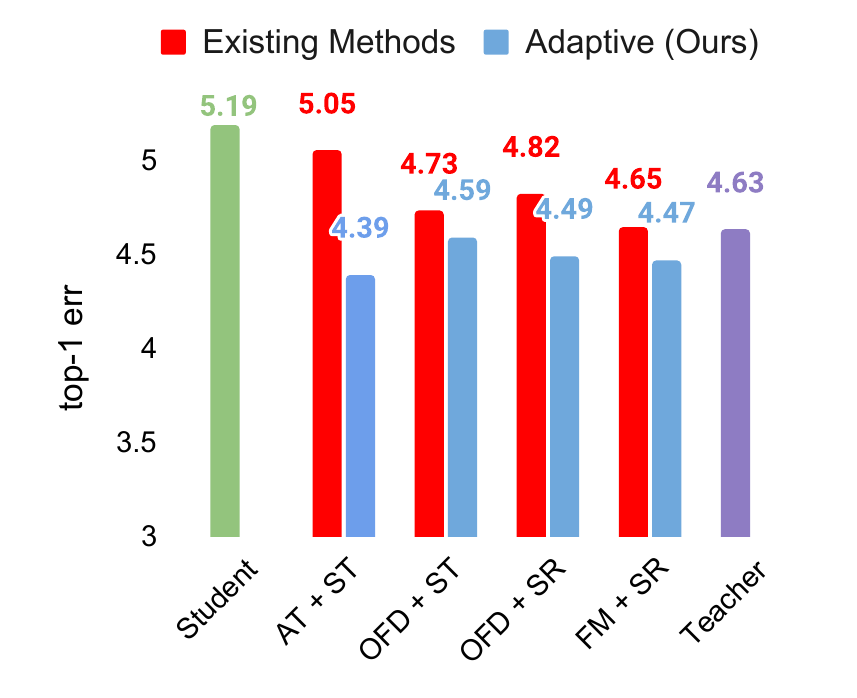}}
    \subfigure[CIFAR-100]{\label{fig:cifar100_sota}\includegraphics[width=0.32\textwidth]{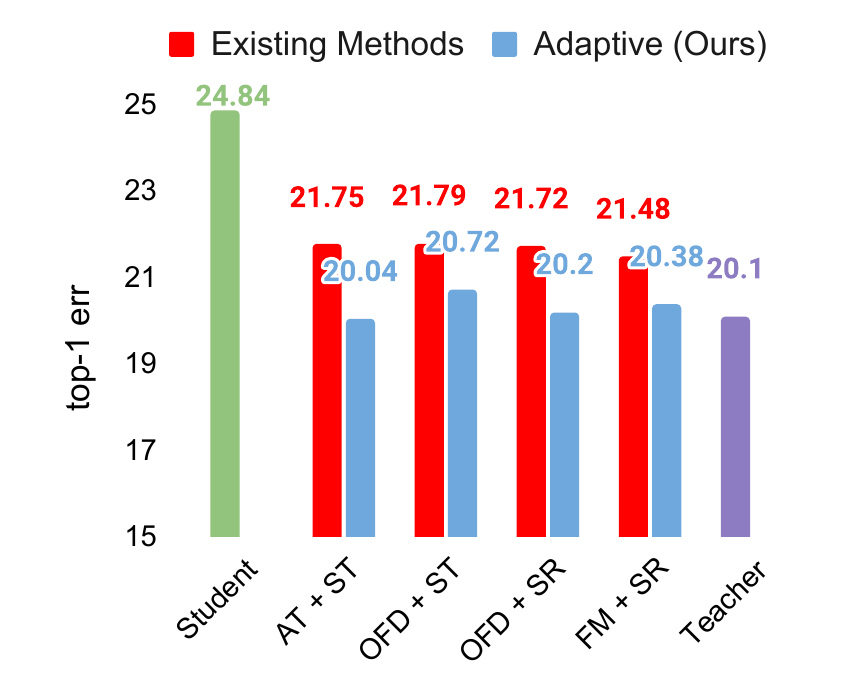}}
    \subfigure[ImageNet-200]{\label{fig:tiny-imagenet_sota}\includegraphics[width=0.32\textwidth]{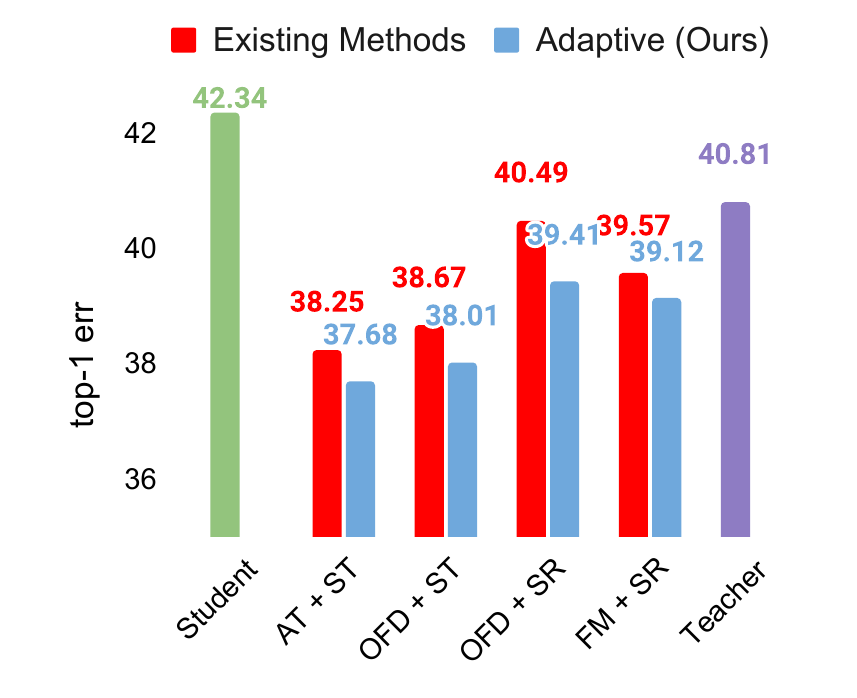}}
    \caption{Comparison between several combinations of existing knowledge distillation methods using Hand-tuned vs Adaptive (ours) methods on (a) CIFAR-10, (b) CIFAR-100 and (c) ImageNet-200 datasets. AT: attention transfer~\cite{zagoruyko2016paying}, ST: soft target~\cite{hinton2015distilling}, OFD: overhaul of feature distillation~\cite{heo2019comprehensive}, FM: feature map~\cite{yang2021knowledge}, and SR: Softmax regression and representation learning~\cite{yang2021knowledge}.}
    \label{fig:image-classification-sota}
\end{figure*}

As it can be seen in Table~\ref{tab:image-classification-results}, we report the performance of teacher and student models without any knowledge distillation, student with single knowledge distillation path using hand-tuned weights, and baselines with multiple knowledge distillation paths which include hand-tuned weights, equal weights, and multiobjective optimization. We observe that adaptive distillation outperforms all other baselines in all three datasets. Among the baseline methods, the multiobjective optimization achieved better results on CIFAR-10 than others while hand-tuned and equal weights achieved better results on CIFAR-100 and ImageNet-200 respectively. Our proposed adaptive distillation method always outperforms the baseline methods and also the teacher model in terms of top1 classification error, as well as top1 agreement error between the teacher and student networks, which demonstrate the efficacy of our proposed aggregation approach for the knowledge distillation process. 

\textbf{Comparisons with state-of-the-arts:}
In Figure~\ref{fig:image-classification-sota}, we study the performance of our adaptive distillation methods when added to a combination of existing state-of-the-art distillation methods for image classification task on CIFAR-10, CIFAR-100 and ImageNet-200 datasets. In addition to Attention Transfer (AT)~\cite{zagoruyko2016paying} and Soft Target (ST)~\cite{hinton2015distilling} distillation methods, we use the recent and advanced distillation methods like Overhaul of Feature Distillation (OFD)~\cite{heo2019comprehensive}, Feature Map (FM)~\cite{yang2021knowledge} and Softmax Regression and Representation Learning (SR)~\cite{yang2021knowledge}. For each combination, we chose two distillation paths that include distillation at intermediate features (AT, OFD, FM) and logits (ST, SR). We observed that adding adaptive methods to the existing methods consistently improved the performance. This experiment shows that our Adaptive Distillation is orthogonal to other knowledge distillation methods and can be used on top of those approaches to boost their performance.

\begin{table*}[!t]
    \centering
    \resizebox{0.9\textwidth}{!}{
        \begin{tabular}{lccccllll}

            \toprule
                               & \multicolumn{4}{c}{\textbf{Path(s)}} & \multicolumn{2}{c}{\textbf{CIFAR-100}} & \multicolumn{2}{c}{\textbf{ImageNet-200}}                                                                                                                                                                       \\
            \cmidrule(r){2-5}\cmidrule(r){6-7}\cmidrule(r){8-9}

                               & \textbf{AT}                          & \textbf{ST}                            & \textbf{NST}                              & \textbf{$\ell_2$-Logit} & \textbf{top1 err} ($\downarrow$) & \textbf{top1 agr} ($\downarrow$) & \textbf{top1 err} ($\downarrow$) & \textbf{top1 agr} ($\downarrow$) \\ \midrule
            Student            &                                      &                                        &                                           &                         & 24.84 \Chartr{1.00}              & -                                & 42.34 \Chartr{1.00}              & -                                \\ \midrule
            Single             & $\checkmark$                         &                                        &                                           &                         & 22.87 \Chartr{0.60}              & 19.04 \Chartb{0.86}              & \textbf{39.80} \Chartr{0.47}     & 35.07 \Chartb{0.75}              \\
            Single             &                                      & $\checkmark$                           &                                           &                         & 22.67 \Chartr{0.56}              & 19.71 \Chartb{1.00}              & 40.24 \Chartr{0.56}              & 35.38 \Chartb{0.78}              \\
            Single             &                                      &                                        & $\checkmark$                              &                         & \textbf{22.45} \Chartr{0.51}     & \textbf{18.61} \Chartb{0.77}     & 40.32 \Chartr{0.58}              & 37.48 \Chartb{1.00}              \\
            Single             &                                      &                                        &                                           & $\checkmark$            & 22.47 \Chartr{0.52}              & 19.18 \Chartb{0.89}              & 39.95 \Chartr{0.50}              & \textbf{34.15} \Chartb{0.66}     \\ \midrule
            Adaptive           & $\checkmark$                         & $\checkmark$                           &                                           &                         & \textbf{20.04} \Chartr{0.0}     & 15.89 \Chartb{0.19}              & \textbf{37.68} \Chartr{0.0}     & 29.46 \Chartb{0.18}              \\
            Adaptive           &                                      &                                        & $\checkmark$                              & $\checkmark$            & 20.49 \Chartr{0.11}              & 15.55 \Chartb{0.12}              & 38.67 \Chartr{0.23}              & 30.71 \Chartb{0.30}              \\
            Adaptive           & $\checkmark$                         & $\checkmark$                           & $\checkmark$                              & $\checkmark$            & 20.16 \Chartr{0.04}              & \textbf{15.09} \Chartb{0.0}     & 38.29 \Chartr{0.15}              & \textbf{27.84} \Chartb{0.0}     \\ \midrule
            Adaptive-layerwise & $\checkmark$                         & $\checkmark$                           &                                           &                         & \textbf{20.43} \Chartr{0.10}     & 15.57 \Chartb{0.12}              & \textbf{38.32} \Chartr{0.16}     & 29.44 \Chartb{0.17}              \\
            Adaptive-layerwise &                                      &                                        & $\checkmark$                              & $\checkmark$            & 20.97 \Chartr{0.21}              & 16.14 \Chartb{0.24}              & 39.50 \Chartr{0.40}              & 30.42 \Chartb{0.28}              \\
            Adaptive-layerwise & $\checkmark$                         & $\checkmark$                           & $\checkmark$                              & $\checkmark$            & 20.85 \Chartr{0.19}              & \textbf{15.25} \Chartb{0.06}     & 38.34 \Chartr{0.16}              & \textbf{28.75} \Chartb{0.10}     \\\midrule
            Teacher            &                                      &                                        &                                           &                         & 20.10 \Chartr{0.03}              & -                                & 40.81 \Chartr{0.68}              & -                                \\ \bottomrule
        \end{tabular}
    }
    \caption{Validation results on CIFAR-100 and ImageNet-200 with additional knowledge distillation paths using neural selective transfer (NST) \cite{huang2017like} and ($\ell_2$-Logit)~\cite{ba2013deep}.}
    \label{tab:ablation-classification}
\end{table*}

\textbf{Ablations:}
In Table \ref{tab:ablation-classification}, we present how adaptive distillation performs when we add more knowledge distillation paths. We add neural selective transfer (NST)~\cite{huang2017like} for feature level distillation and regression logits ($\ell_2$-Logit)~\cite{ba2013deep} for output layers in addition to existing paths. We use $\bm{v}$ = $[v_{AT}, v_{ST}, v_{NST}, v_{\ell_2-Logit}]$ = $[1000, 0.1, 10, 0.1]$ for hand-tuned baselines. We construct three variants of adaptive distillation methods by choosing [AT, ST], [NST, $\ell_2$-Logit] and [AT, ST, NST, $\ell_2$-Logit]. Finally, we also explore the robustness of adaptive distillation by treating each of the 4 residual layers in the backbone as a unique path for distillation. This results in 5 (4+1) paths for [AT, ST] and [NST, $\ell_2$-Logit] experiments, and 10 (8+2) paths for [AT, ST, NST, $\ell_2$-Logit]. We refer to these models as `Adaptive-layerwise'. For more information on different structures refer to Appendix~\ref{app:exp}.

\begin{table*}[!htb]
        \centering
        \addtolength{\tabcolsep}{-4pt}
        \resizebox{.82\textwidth}{!}{
            \begin{tabular}{lcccccllll}

                \toprule
                                                        & & \multicolumn{4}{c}{\textbf{Path(s)}}   &
                             \multicolumn{2}{c}{\textbf{CIFAR-100}} & \multicolumn{2}{c}{\textbf{ImageNet-200}}                                                                                                                                                                                                     \\
                \cmidrule(r){3-6}\cmidrule(r){7-8}\cmidrule(r){9-10}

                                                        & \textbf{$\alpha$}                      & \textbf{AT}                               & \textbf{ST}  & \textbf{NST} & \textbf{$\ell_2$-Logit} & \textbf{top1 err} ($\downarrow$) & \textbf{top1 agr} ($\downarrow$) & \textbf{top1 err} ($\downarrow$) & \textbf{top1 agr} ($\downarrow$) \\ \midrule
                \multicolumn{2}{l}{Student (ResNet-18)} &                                        &                                           &              &              & 24.84 \Chartr{1.00}     & \multicolumn{1}{c}{-}            & 42.34 \Chartr{1.00}              & \multicolumn{1}{c}{-}                                               \\ \midrule
                Adaptive                                & 0.25                                   & $\checkmark$                              & $\checkmark$ &              &                         & 21.14 \Chartr{0.26}              & 16.97  \Chartb{0.82}             & 39.13  \Chartr{0.33}             & 33.53 \Chartb{0.87}              \\
                Adaptive                                & 0.50                                   & $\checkmark$                              & $\checkmark$ &              &                         & 20.42 \Chartr{0.12}              & 16.33  \Chartb{0.56}             & 39.56  \Chartr{0.42}             & 32.03 \Chartb{0.65}              \\
                Adaptive                                & 0.75                                   & $\checkmark$                              & $\checkmark$ &              &                         & 20.80 \Chartr{0.19}              & 16.19  \Chartb{0.50}             & 38.50  \Chartr{0.19}             & 30.55 \Chartb{0.42}              \\
                Adaptive                                & 1.00                                   & $\checkmark$                              & $\checkmark$ &              &                         & \textbf{20.04} \Chartr{0.04}     & \textbf{15.89} \Chartb{0.37}     & \textbf{37.68} \Chartr{0.02}     & \textbf{29.46} \Chartb{0.26}     \\ \midrule
                Adaptive                                & 0.25                                   &                                           &              & $\checkmark$ & $\checkmark$            & 21.40 \Chartr{0.31}              & 17.40  \Chartb{1.00}             & 39.05  \Chartr{0.31}             & 34.39 \Chartb{1.00}              \\
                Adaptive                                & 0.50                                   &                                           &              & $\checkmark$ & $\checkmark$            & 21.08 \Chartr{0.25}              & 16.74  \Chartb{0.73}             & 39.18  \Chartr{0.34}             & 32.57 \Chartb{0.73}              \\
                Adaptive                                & 0.75                                   &                                           &              & $\checkmark$ & $\checkmark$            & \textbf{20.24} \Chartr{0.08}     & \textbf{15.36} \Chartb{0.15}     & \textbf{38.40} \Chartr{0.17}     & \textbf{30.70} \Chartb{0.45}     \\
                Adaptive                                & 1.00                                   &                                           &              & $\checkmark$ & $\checkmark$            & 20.49 \Chartr{0.13}              & 15.55  \Chartb{0.23}             & 38.67  \Chartr{0.23}             & 30.71 \Chartb{0.45}              \\ \midrule
                Adaptive                                & 0.25                                   & $\checkmark$                              & $\checkmark$ & $\checkmark$ & $\checkmark$            & 20.28 \Chartr{0.09}              & 16.11  \Chartb{0.46}             & 38.34  \Chartr{0.16}             & 31.72 \Chartb{0.60}              \\
                Adaptive                                & 0.50                                   & $\checkmark$                              & $\checkmark$ & $\checkmark$ & $\checkmark$            & 20.46 \Chartr{0.12}              & 15.92  \Chartb{0.39}             & 38.02  \Chartr{0.09}             & 29.69 \Chartb{0.29}              \\
                Adaptive                                & 0.75                                   & $\checkmark$                              & $\checkmark$ & $\checkmark$ & $\checkmark$            & \textbf{19.94} \Chartr{0.0}      & 15.34  \Chartb{0.15}             & \textbf{37.84} \Chartr{0.05}     & 28.69 \Chartb{0.14}              \\
                Adaptive                                & 1.00                                   & $\checkmark$                              & $\checkmark$ & $\checkmark$ & $\checkmark$            & 20.16 \Chartr{0.06}              & \textbf{15.09} \Chartb{0.04}     & 38.29          \Chartr{0.15}     & \textbf{27.84} \Chartb{0.02}     \\ \midrule
                \multicolumn{2}{l}{Teacher (ResNet-50)} &                                        &                                           &              &              & 20.10 \Chartr{0.04}     & \multicolumn{1}{c}{-}            & 40.81 \Chartr{0.68}              & \multicolumn{1}{c}{-}                                               \\ \bottomrule
            \end{tabular}
        }
        \caption{Validation results on CIFAR-100 and ImageNet-200 by varying $\alpha$ (the weight indicating the importance of the distillation loss compared to the main empirical loss).  When the number of distillation paths increases, the lower $\alpha$ values are preferable.}
        \label{tab:ablation-classification-alphas-1}
    \end{table*}

       \begin{table*}[!t]
        \addtolength{\tabcolsep}{-4pt}
        \centering
        \resizebox{.85\textwidth}{!}{
            \begin{tabular}{lcccccllll}

                \toprule
                                    & & \multicolumn{4}{c}{\textbf{Path(s)}} & \multicolumn{2}{c}{\textbf{CIFAR-100}} & \multicolumn{2}{c}{\textbf{ImageNet-200}}                                                                                                                                                                                      \\
                \cmidrule(r){3-6}\cmidrule(r){7-8}\cmidrule(r){9-10}

                                    & \textbf{$\alpha$}                    & \textbf{AT}                            & \textbf{ST}                               & \textbf{NST} & \textbf{$\ell_2$-Logit} & \textbf{top1 err} ($\downarrow$) & \textbf{top1 agr} ($\downarrow$) & \textbf{top1 err} ($\downarrow$) & \textbf{top1 agr} ($\downarrow$) \\ \midrule
                Student (ResNet-10) &                                      &                                        &                                           &              &                         & 27.06  \Chartr{1.00}             & -                                & 45.75 \Chartr{1.00}              & -                                \\ \midrule
                Single              & 1.00                                 & $\checkmark$                           &                                           &              &                         & 26.29  \Chartr{0.89}             & 23.43 \Chartb{1.00}              & \textbf{43.89}  \Chartr{0.63}    & \textbf{37.69} \Chartb{0.68}     \\
                Single              & 1.00                                 &                                        & $\checkmark$                              &              &                         & 25.31  \Chartr{0.75}             & 22.82 \Chartb{0.86}              & 44.23 \Chartr{0.70}              & 38.35          \Chartb{0.77}     \\
                Single              & 1.00                                 &                                        &                                           & $\checkmark$ &                         & \textbf{24.61} \Chartr{0.65}     & \textbf{21.53} \Chartb{0.55}     & 43.98 \Chartr{0.65}              & 39.86          \Chartb{1.00}     \\
                Single              & 1.00                                 &                                        &                                           &              & $\checkmark$            & 24.76  \Chartr{0.67}             & 22.16 \Chartb{0.70}              & 44.65 \Chartr{0.78}              & 37.90          \Chartb{0.71}     \\ \midrule
                Adaptive            & 1.00                                 & $\checkmark$                           & $\checkmark$                              &              &                         & 23.88  \Chartr{0.55}             & 19.78 \Chartb{0.14}              & 41.36 \Chartr{0.13}              & 34.76          \Chartb{0.24}     \\
                Adaptive            & 0.75                                 &                                        &                                           & $\checkmark$ & $\checkmark$            & 23.13  \Chartr{0.44}             & 19.64 \Chartb{0.11}              & 43.15 \Chartr{0.48}              & 36.26          \Chartb{0.46}     \\
                Adaptive            & 0.75                                 & $\checkmark$                           & $\checkmark$                              & $\checkmark$ & $\checkmark$            & \textbf{22.77}  \Chartr{0.39}             & \textbf{19.27} \Chartb{0.02}              & \textbf{40.93} \Chartr{0.04}              & \textbf{33.25}          \Chartb{0.01}     \\ \midrule
                Teacher (ResNet-50) &                                      &                                        &                                           &              &                         & 20.10  \Chartr{0.01}             & -                                & 40.81 \Chartr{0.02}              & -                                \\ \bottomrule
            \end{tabular}
        }
        \caption{Validation results on CIFAR-100 and ImageNet-200 for Student ResNet10 and Teacher ResNet50. The effect of number of distillation paths can be seen in this table since layerwise approaches have more paths and need smaller $\alpha$ values to improve the performance of the student model.}
        \label{tab:student-resnet10}
    \end{table*}

We observed that our proposed adaptive methods continue to outperform their single baselines in terms of both top1 classification error and top1 agreement error. Adaptive methods with [AT, ST] paths yielded the best results in terms of top1 classification error compared to other counterparts. However, the adaptive method with [AT, ST, NST, $\ell_2$-Logit] achieves the best agreement error, which is the objective of the distillation part. This suggests that shifting the focus from distillation (i.e, reducing the value of $\alpha$) to main loss could improve top1 classification error. Also, our adaptive methods with [AT, ST, NST, $\ell_2$-Logit] paths performed slightly better than [NST, $\ell_2$-Logit] and worse compared to [AT, ST] paths. This suggests that adaptive distillation can reduce negative effects caused by [NST, $\ell_2$-Logit] to some degree. The same patterns can be observed with layerwise approaches as well.

\begin{table*}[!tb]
    \addtolength{\tabcolsep}{-5.5pt}
    \centering
    \resizebox{0.6\textwidth}{!}{%
\begin{tabular}{lccllll}
    \toprule
                      & \multicolumn{2}{c}{\textbf{Path(s)}} & \multicolumn{2}{c}{\textbf{Cityscapes}} & \multicolumn{2}{c}{\textbf{ADE 20K}}                                                                                                                                                             \\ 
\cmidrule(r){2-3}\cmidrule(r){4-5}\cmidrule(r){6-7}
                      & \textbf{AT}                          & \textbf{ST}                             & \multicolumn{1}{c}{\textbf{mIoU} ($\uparrow$)} & \multicolumn{1}{c}{\textbf{mAcc ($\uparrow$)}} &  \multicolumn{1}{c}{\textbf{mIoU}($\uparrow$)} & \multicolumn{1}{c}{\textbf{mAcc} ($\uparrow$)} \\ \midrule
    Student           &                                      &                                         & \hspace{5pt} 73.42 \Chartg{0.041}              & \hspace{5pt} 81.33 \Chartb{0.68}              & \hspace{15pt} 32.84 \Chartg{0.09}             & \hspace{5pt} 42.50 \Chartb{0.09}                   \\ \midrule
    Single            & $\checkmark$                         &                                         & \hspace{5pt} \textbf{75.75} \Chartg{1.000}     & \hspace{5pt} \textbf{83.59} \Chartb{1.00}              & \hspace{15pt} 34.03 \Chartg{0.39}             & \hspace{5pt} 43.69 \Chartb{0.33}                        \\
    Single            &                                      & $\checkmark$                            & \hspace{5pt} 75.25 \Chartg{0.794}              & \hspace{5pt} 82.89 \Chartb{0.90}              & \hspace{15pt} 33.19 \Chartg{0.18}             & \hspace{5pt} 42.79 \Chartb{0.15}                        \\
    Hand-tuned        & $\checkmark$                         & $\checkmark$                            & \hspace{5pt} 75.67 \Chartg{0.967}              & \hspace{5pt} 83.49 \Chartb{0.99}              & \hspace{15pt} 33.43 \Chartg{0.24}             & \hspace{5pt} 43.66 \Chartb{0.33}                        \\
    Equal             & $\checkmark$                         & $\checkmark$                            & \hspace{5pt} 74.38 \Chartg{0.436}              & \hspace{5pt} 81.77 \Chartb{0.74}              & \hspace{15pt} 34.06 \Chartg{0.40}             & \hspace{5pt} 42.95 \Chartb{0.18}                        \\
    Multiobjective    & $\checkmark$                         & $\checkmark$                            & \hspace{5pt} 68.99 \Chartg{0.041}              & \hspace{5pt} 77.19 \Chartb{0.09}              & \hspace{15pt} 34.01 \Chartg{0.39}             & \hspace{5pt} 43.04 \Chartb{0.20}                        \\
    \textbf{Adaptive} & $\checkmark$                         & $\checkmark$                            & \hspace{5pt} 75.43 \Chartg{0.868}              & \hspace{5pt} 82.51 \Chartb{0.85}              & \hspace{15pt} \textbf{34.73} \Chartg{0.57}             & \hspace{5pt} \textbf{44.28} \Chartb{0.45}                        \\ \midrule
    Teacher           &                                      &                                         & \hspace{5pt} 74.52 \Chartg{0.494}              & \hspace{5pt} 81.86 \Chartb{0.75}              & \hspace{15pt} 36.45 \Chartg{1.00}             & \hspace{5pt} 46.96 \Chartb{1.00}                  \\ \bottomrule
    \end{tabular}
    }
    \caption{Performance of knowledge distillation methods with multiple paths on Cityscapes \cite{Cordts2016Cityscapes} and ADE 20K \cite{Zhou_2017_CVPR} datasets  for semantic segmentation.$(\uparrow)$ indicates higher the better.}
    \label{tab:segmentation}
\end{table*}

\begin{table*}[!tb]
    \addtolength{\tabcolsep}{-4.5pt}
    \centering
    \resizebox{0.74\textwidth}{!}{

\begin{tabular}{lccllccll}
    \toprule
                             & \multicolumn{2}{c}{\textbf{Path(s)}} & \multicolumn{1}{c}{\textbf{Cityscapes}} & \multicolumn{1}{c}{\textbf{COCO}}             & \multicolumn{2}{c}{\textbf{Path(s)}}          & \multicolumn{1}{c}{\textbf{Cityscapes}} & \multicolumn{1}{c}{\textbf{COCO}}                                                                                                 \\
    \cmidrule(r){2-3}\cmidrule(r){4-4}\cmidrule(r){5-5}\cmidrule(r){6-7}\cmidrule(r){8-8}\cmidrule(r){9-9}

                             & \textbf{Feats\textsubscript{B}}      & \textbf{Box}                            & \multicolumn{1}{c}{\textbf{mAP} ($\uparrow$)} & \multicolumn{1}{c}{\textbf{mAP} ($\uparrow$)} & \textbf{Feats\textsubscript{P}}         & \textbf{Box}                      & \multicolumn{1}{c}{\textbf{mAP} ($\uparrow$)} & \multicolumn{1}{c}{\textbf{mAP} ($\uparrow$)} \\ \midrule
    Student                  &                                      &                                         & \hspace{15pt} 38.1 \Chartg{0.02}              & 36.1 \Chartb{0.09}                                     &                                         &                                   & \hspace{15pt} 38.1 \Chartg{0.02}              & 36.1 \Chartb{0.09}                                \\\midrule
    Single                   & $\checkmark$                         &                                         & \hspace{15pt} 39.6 \Chartg{0.33}              & 36.3 \Chartb{0.12}                                     & $\checkmark$                            &                                   & \hspace{15pt} 39.9 \Chartg{0.39}              & 36.8 \Chartb{0.19}                                 \\
    Single                   &                                      & $\checkmark$                            & \hspace{15pt} 39.8 \Chartg{0.37}              & 36.2 \Chartb{0.11}                                     &                                         & $\checkmark$                      & \hspace{15pt} 39.8 \Chartg{0.37}              & 36.3 \Chartb{0.12}                                 \\
    Hand-tuned               & $\checkmark$                         & $\checkmark$                            & \hspace{15pt} 39.7 \Chartg{0.35}              & 36.4 \Chartb{0.13}                                     & $\checkmark$                            & $\checkmark$                      & \hspace{15pt} 40.1 \Chartg{0.43}              & 36.7 \Chartb{0.17}                                 \\
    Multiobjective           & $\checkmark$                         & $\checkmark$                            & \hspace{15pt} 40.0 \Chartg{0.41}              & 36.5 \Chartb{0.15}                                     & $\checkmark$                            & $\checkmark$                      & \hspace{15pt} 40.6 \Chartg{0.53}              & 36.8 \Chartb{0.19}                                 \\
    \textbf{Adaptive}        & $\checkmark$                         & $\checkmark$                            & \hspace{15pt} \textbf{40.2} \Chartg{0.45}     & \textbf{37.1} \Chartb{0.23}                                     & $\checkmark$                            & $\checkmark$                      & \hspace{15pt} \textbf{40.9} \Chartg{0.59}     & \textbf{37.7} \Chartb{0.31}                                 \\ \midrule
    Teacher       &                                      &                                         & \hspace{15pt} 42.9 \Chartg{1.00}              & 42.7 \Chartb{1.00}                                     &                                         &                                   & \hspace{15pt} 42.9 \Chartg{1.00}              & 42.7 \Chartb{1.00}                                \\ \bottomrule
\end{tabular}
    }
    \caption{Performance of knowledge distillation with multiple paths on Cityscapes~\cite{Cordts2016Cityscapes} and COCO~\cite{lin2014microsoft} datasets for object detection.$(\uparrow)$ indicates higher the better. Feats\textsubscript{B}: features from backbone layers, Feats\textsubscript{P}: features from pyramid layers, and Box: bounding box generator.}
    \label{tab:cityscapes}
\end{table*}

\textbf{Importance of distillation loss ($\alpha$)} As it was mentioned in Section~\ref{sec:exp}, when the number of distillation paths increases, the alignment between teacher and student increases (the agreement error decreases), but the performance of the student model does not necessarily improve. This is mainly due to the higher importance of the distillation when the number of paths increases. To remedy this, we suggest to decrease the weight of distillation using its parameter $\alpha$. In Table~\ref{tab:ablation-classification-alphas-1}, we change $\alpha$, the relative importance of distillation loss compared to main empirical loss. We observed that higher value of alphas led to lower top1 agreement error between teacher and student networks. We also observed that as the number of distillation paths are increased, the norm of the gradients from distillation loss increases, which may improve top1 agreement error but might not always improve the top1 classification error. A relatively lower $\alpha$, reduces the weight of distillation gradients and bolster to improve top1 classification error when more distillation paths are used. For example, changing $\alpha$ to 0.75 instead of 1.0 helped improve the top1 classification error when all [AT, ST, NST and $\ell_2$-Logit] distillation paths were used on CIFAR-100 and ImageNet-200 datasets. We observe a similar trend with the layerwise methods in Table~\ref{tab:ablation-classification-alphas-2}. On the other hand, it can be seen that using lower $\alpha$ values increase both top1 agreement and top1 classification errors.

\textbf{Smaller Student Models:} In Table~\ref{tab:student-resnet10}, we experiment with smaller ResNet model with only 10 layers. We observe that adaptive distillation with multiple paths are outperforming their single distillation counterparts even when the student model is smaller and the gap between teacher and student is higher.

\textbf{Semantic Segmentation:}
For this task, we perform experiments on Cityscapes~\cite{Cordts2016Cityscapes} and ADE20K \cite{Zhou_2017_CVPR} datasets and evaluate mean intersection over union (mIoU) and mean accuracy (mAcc). We use semantic fpn~\cite{kirillov2019panoptic} network with ResNet~\cite{he2016deep} backbone(s). We initialize our backbone(s) with ImageNet pretrained weights. We use attention transfer (AT) for backbone features and soft target (ST) for outputs. We flatten the semantic segmentation output to generate soft targets. We chose an SGD optimizer with an initial learning rate of 0.01, polynomial learning rate policy (power=0.9) and train for 80K iterations with a batch size 8 (512x1024 crop) images.  Table~\ref{tab:segmentation} indicates the performance of knowledge distillation with multiple paths for this task. We observe that whenever the gap between teacher and student models' performance is large, adaptive distillation outperforms other approaches in this task. This can be inferred from results in ADE 20K dataset, where adaptive distillation outperforms hand-tuned ones. When this gap is small, adaptive distillation can achieve very close performance to the best hand-tuned model, which essentially reduces the time for exhaustive search for best hyperparameters.

\textbf{Object Detection:} We perform experiments on Cityscapes~\cite{Cordts2016Cityscapes} and COCO \cite{lin2014microsoft} datasets and evaluate mean average precision(mAP). We use RetinaNet~\cite{lin2017focal} with Generalized Focal Loss~\cite{li2020generalized} built using ResNet~\cite{he2016deep} backbone(s). We use ResNet18 as student backbone and ResNet101 as the teacher backbone. We use feature-based knowledge distillation~\cite{zhang2021improve} for features (`Feats\textsubscript{B}' indicate features from the backbone layers and `Feats\textsubscript{P}' indicate features from the pyramid layers) and localization distillation \cite{zheng2021LD} for bounding box generator (referred as `Box'). We initialize our backbone(s) with coco pretrained weights. We chose an SGD optimizer with an initial learning rate of (0.01, 0.02) and train for a total of (64, 12) epochs on Cityscapes and coco datasets. We reduce the learning rate by factor of 10 after 56 epochs on Cityscapes while after 8, 11 epochs on coco dataset. We use a batch size 8 (512x1024 crop) images. For hand-tuned weights, we use the hyperparameters suggested by the respective authors. Results on the Table \ref{tab:cityscapes} indicate that adaptive distillation and multiobjective outperform Hand-tuned baselines. When `Feats\textsubscript{B}' were used for feature distillation along with Bbox distillation, hand-tuned baseline failed to minimize the negative effects caused by distillation on backbone features.

%% file: 6-conclusion.tex
\section{Conclusion and Future Directions}\label{sec:con}
In this paper, we explored different methods for aggregating different paths for efficient knowledge distillation from a teacher to a student. We proposed an adaptive approach, with which we intend to learn the importance of each distillation path during the training process. The effectiveness of this approach is being corroborated by our extensive empirical studies on various tasks such as classification, semantic segmentation, and object detection. Moreover, we introduce another baseline for this problem based on multiobjective optimization.

Although our approach has been examined on a single teacher with multiple distillation paths, the extension of these methods to multiple teacher can be investigated in future works. Moreover, as another future direction, a theoretical investigation of these approaches can further illuminate the effect of each path on the distillation process. 

%% file: 7-appendix.tex
\section{Additional Experimental Details}~\label{app:exp}
In this section, we will provide more details regarding our approach and the empirical studies discussed in the main body. In Section~\ref{sec:exp}, we present results of applying different methods for distilling the knowledge from different paths.

\subsection{Distillation Losses}
In this part, we explain each knowledge distillation paths we used in our experimental studies.
\paragraph{Soft Target} First, we start by the soft target (ST)~\cite{hinton2015distilling}, which is the primary form of distillation used in many different applications. In this form of distillation the goal is for the student model to match the probability distribution of the teacher for each data sample. Hence, if the input data is $\bm{x}$, by denoting the soft target of each model as:
$$q_i\left(\bm{x}|\tau\right) = \frac{e^{\frac{z_i\left(\bm{x}\right)}{\tau}}}{\sum_{j \in [C]} e^{\frac{z_j\left(\bm{x}\right)}{\tau}}},\;\;\; \forall i\in[C],$$ where $\tau$ is the temperature for the soft targets and $z_i\left(.\right)$s are the logits for each class generated by the model. The total number of classes is $C$. Thus, the loss for this distillation path is the cross entropy between the soft targets of the teacher and the student defined as:
\begin{equation*}\label{eq:ST}
    \mathcal{L}_{ST}\left(\bm{x};\bm{w}^S,\bm{w}^T\right) = -\tau^2 \sum_{i\in[C]} q_i^T\left(\bm{x}|\tau\right)\log{q_i^S\left(\bm{x}|\tau\right)},
\end{equation*}

where $q_i^T$ and $q_i^S$ are the soft targets of the student and the teacher, respectively.

\paragraph{Attention Transfer} This type of distillation is used on hint layer features and tries to minimize the distance between the feature maps of the student and teacher as defined in~\cite{zagoruyko2016paying}. The loss function for this distillation path can be written as:
\begin{equation*}
    \mathcal{L}_{A T}\left(\bm{x};\bm{w}^S,\bm{w}^T\right)= \sum_{j \in [N]}\left\|\frac{\bm{a}_{j}^{S}}{\left\|\bm{a}_{j}^{S}\right\|_{2}}-\frac{\bm{a}_{j}^{T}}{\left\|\bm{a}_{j}^{T}\right\|_{2}}\right\|_{2}
    \label{eq:at}
\end{equation*}
where $\bm{a}_{j}^{S}=\operatorname{vec}\left(\bm{A}_{j}^{S}\right)$ and $\bm{a}_{j}^{T}=\operatorname{vec}\left(\bm{A}_{j}^{T}\right)$ are respectively the $j$-th pair of student and teacher attention maps in vectorized form, and $p$ refers to norm type. The attention maps $\bm{A}_j \in \mathbb{R}^{H_j\times W_j}$ are computed by adding the transformed features across its channel dimension $D_j$ of feature map $\mathcal{F}_j \in \mathbb{R}^{H_j\times W_j \times D_j}$ for both student and teacher models:
\begin{equation*}
    \bm{A}_j = \sum_{d=1}^{D_j}\left(\mathcal{F}_{j,d}\right)^{2}
\end{equation*}

\paragraph{Neural Selective Transfer}
 In this distillation, we match the features (in the spatial dimension) between teacher and student networks by minimizing a special case ($d=1$, $c=0$) of  Maximum Mean Discrepancy (MMD) distance described in~\cite{huang2017like} as,
\begin{equation*}
    \mathcal{L}_{NS T}\left(\bm{x};\bm{w}^S,\bm{w}^T\right)=\left\| \mathbf{G}\left[\mathcal{F}_{j}^{T}\right] - \mathbf{G}\left[\bm{w}_{j}^{A}\mathcal{F}_{j}^{S}\right]\right\|_2
\end{equation*}
where $\mathcal{F}_{j}^T$ and $\bm{w}_{j}^{A}\mathcal{F}_{j}^S$ are $j$-th pair of teacher and adapted student feature maps normalized across the channel dimension and $\mathbf{G}$ is the Gram matrix computed as:
\begin{equation*}
    \mathbf{G}(\mathcal{F}_{D_j\times H_jW_j})=\mathbf{\mathcal{F}}_{D_j\times H_jW_j}^{\top} \mathbf{\mathcal{F}}_{D_j\times H_jW_j}
\end{equation*}
Compared to attention transfer~\cite{zagoruyko2016paying} which computes spatial attention ($\bm{A}_j \in \mathbb{R}^{H_j\times W_j}$) of the feature maps, neural selection transfer~\cite{huang2017like} computes Gram Matrix ($\mathbf{G}_j \in \mathbb{R}^{H_jW_j \times H_jW_j}$) to capture interactions across the spatial domain.

\paragraph{$\mathbf{\ell_2}$-Logit}
This form of distillation is similar to the soft target, but in here, we simply measure the euclidean distance between teacher and student output logits ($\bm{z}^T, \bm{z}^S$) instead of their cross entropy~\cite{ba2013deep}.
\begin{equation*}
    \mathcal{L}_{\ell_2-Logit}\left(\bm{x};\bm{w}^S,\bm{w}^T\right)=\left\| \bm{z}^T\left(\bm{x};\bm{w}^T\right) - \bm{z}^S\left(\bm{x};\bm{w}^S\right) \right\|_2
\end{equation*}

\subsection{Metrics}
For the experimental results in addition to the main metric of the task (classification error for the classification, IoU for the semantic segmentation, and mAP for the object detection tasks), we used top1 agreement error, a metric introduced by~\cite{stanton2021does} to evaluate the performance of knowledge distillation approaches. We compute top1 agreement error using prediction probabilities ($\bm{p}^T \in \mathbb{R}^C$ and $\bm{p}^S \in \mathbb{R}^C$) from teacher and student as:
\begin{align}\label{eq:agreement}
&\text{Top1\textsubscript{Agreement Error}}= \\ \nonumber
&\quad\left\{1 - \frac{1}{N} \sum_{i=1}^{N}[\arg\max_C (\bm{p}_i^T)=\arg\max_C(\bm{p}_i^S)]\right\}\times100
\end{align}

Looking into Eq.~(\ref{eq:agreement}) it can be inferred that this metric wants to evaluate the objective of the primary distillation form as in soft target in Eq.~(\ref{eq:agreement}). As it can be seen in the experimental results, having a better alignment with the teacher does not necessary reflects as a better generalization performance for the student. Especially, in cases when the student can have a better performance than the teacher using knowledge distillation.

\begin{figure*}[!t]
    \centering
    \subfigure[Aggregated loss from different layers of backbone.]{\label{fig:block_wise}\includegraphics[width=110mm]{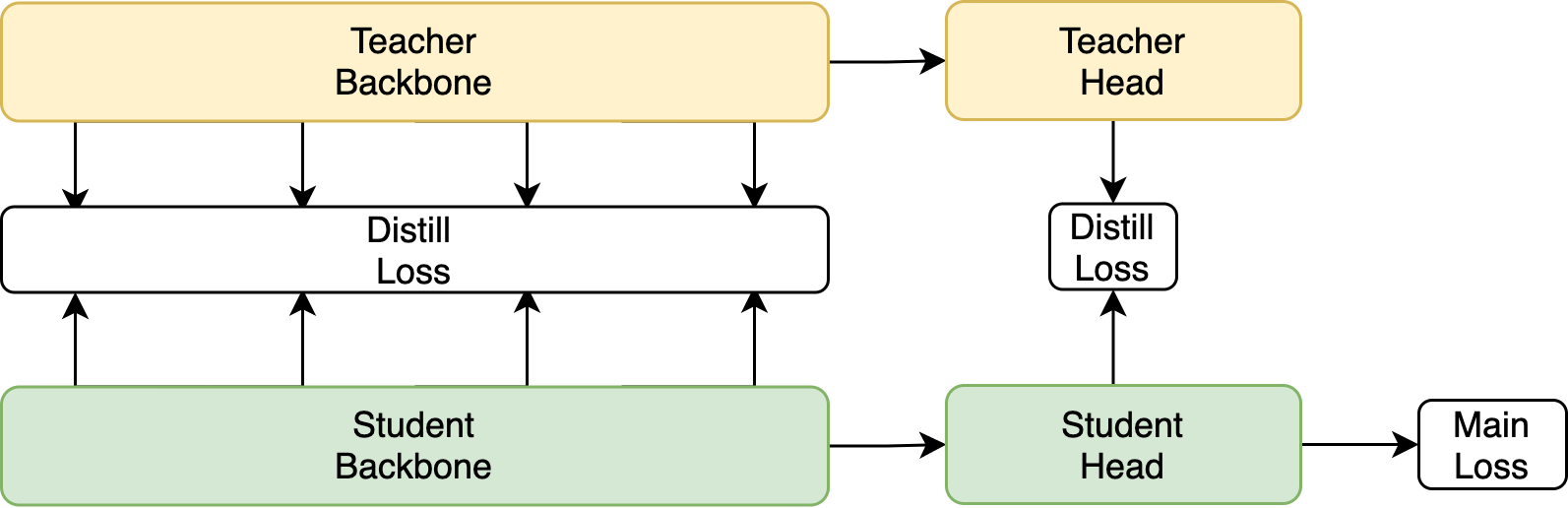}} \hspace{3mm}
    \subfigure[Independent loss from each layer, aka layerwise.]{\label{fig:layer_wise}\includegraphics[width=110mm]{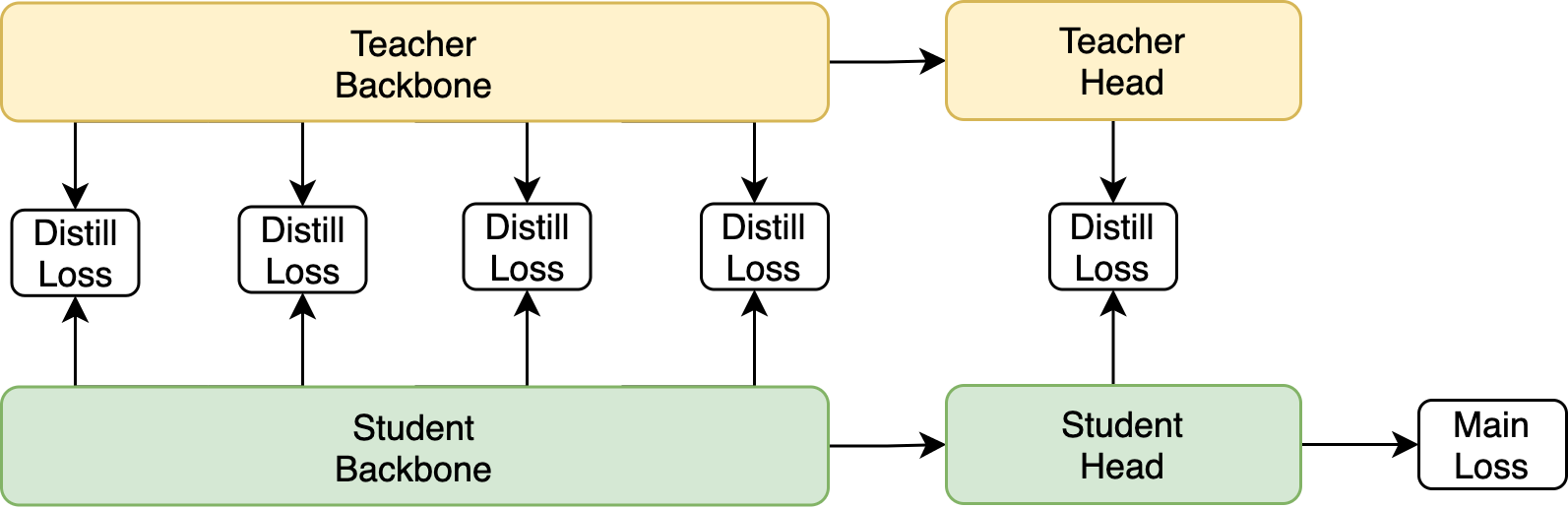}}
    \caption{Knowledge distillation with multiple paths. (a) Backbone features from different layers are passed through a single distillation loss function. (b) Each feature from a backbone layer is passed through a distillation loss function.}
    \label{fig:kd_multipath}
\end{figure*}

\begin{table*}[!t]
    \addtolength{\tabcolsep}{-4pt}
    \centering
    \resizebox{.85\textwidth}{!}{
        \begin{tabular}{lcccccllll}

            \toprule
                                                    & & \multicolumn{4}{c}{\textbf{Path(s)}}   &
                     \multicolumn{2}{c}{\textbf{CIFAR-100}} & \multicolumn{2}{c}{\textbf{ImageNet-200}}                                                                                                                                                                                                     \\
            \cmidrule(r){3-6}\cmidrule(r){7-8}\cmidrule(r){9-10}

                                                    & \textbf{$\alpha$}                      & \textbf{AT}                               & \textbf{ST}  & \textbf{NST} & \textbf{$\ell_2$-Logit} & \textbf{top1 err} ($\downarrow$) & \textbf{top1 agr} ($\downarrow$) & \textbf{top1 err} ($\downarrow$) & \textbf{top1 agr} ($\downarrow$) \\ \midrule
            \multicolumn{2}{l}{Student (ResNet-18)} &                                        &                                           &              &              & 24.84  \Chartr{1.00}    & \multicolumn{1}{c}{-}            & 42.34  \Chartr{1.00}             & \multicolumn{1}{c}{-}                                               \\ \midrule
            Adaptive                                & 0.25                                   & $\checkmark$                              & $\checkmark$ & $\checkmark$ & $\checkmark$            & 20.28  \Chartr{0.09}             & 16.11  \Chartb{1.00}             & 38.34 \Chartr{0.13}              & 31.72 \Chartb{1.00}              \\
            Adaptive                                & 0.50                                   & $\checkmark$                              & $\checkmark$ & $\checkmark$ & $\checkmark$            & 20.46  \Chartr{0.12}             & 15.92  \Chartb{0.84}             & 38.02 \Chartr{0.06}              & 29.69 \Chartb{0.49}              \\
            Adaptive                                & 0.75                                   & $\checkmark$                              & $\checkmark$ & $\checkmark$ & $\checkmark$            & \textbf{19.94} \Chartr{0.02}     & 15.34  \Chartb{0.35}             & \textbf{37.84} \Chartr{0.02}     & 28.69 \Chartb{0.24}              \\
            Adaptive                                & 1.00                                   & $\checkmark$                              & $\checkmark$ & $\checkmark$ & $\checkmark$            & 20.16  \Chartr{0.06}             & \textbf{15.09} \Chartb{0.14}     & 38.29  \Chartr{0.12}             & \textbf{27.84} \Chartb{0.03}     \\ \midrule
            Adaptive-layerwise                      & 0.25                                   & $\checkmark$                              & $\checkmark$ & $\checkmark$ & $\checkmark$            & \textbf{20.21} \Chartr{0.07}     & 15.28  \Chartb{0.30}             & 38.63  \Chartr{0.19}             & 31.58 \Chartb{0.96}              \\
            Adaptive-layerwise                      & 0.50                                   & $\checkmark$                              & $\checkmark$ & $\checkmark$ & $\checkmark$            & 20.89  \Chartr{0.21}             & 15.44  \Chartb{0.44}             & \textbf{38.07}  \Chartr{0.07}    & 29.89 \Chartb{0.54}              \\
            Adaptive-layerwise                      & 0.75                                   & $\checkmark$                              & $\checkmark$ & $\checkmark$ & $\checkmark$            & 20.55  \Chartr{0.14}             & \textbf{15.02} \Chartb{0.08}     & 38.80  \Chartr{0.23}             & \textbf{28.39} \Chartb{0.16}     \\
            Adaptive-layerwise                      & 1.00                                   & $\checkmark$                              & $\checkmark$ & $\checkmark$ & $\checkmark$            & 20.85  \Chartr{0.20}             & 15.25  \Chartb{0.28}             & 38.34 \Chartr{0.13}              & 28.75 \Chartb{0.25}              \\ \midrule
            \multicolumn{2}{l}{Teacher (ResNet-50)} &                                        &                                           &              &              & 20.10  \Chartr{0.05}    & \multicolumn{1}{c}{-}            & 40.81  \Chartr{0.67}             & \multicolumn{1}{c}{-}                                               \\ \bottomrule
        \end{tabular}
    }
    \caption{Validation results on CIFAR-100 and ImageNet-200 by varying $\alpha$ (the weight indicating the importance of the distillation loss compared to the main empirical loss).}
    \label{tab:ablation-classification-alphas-2}
\end{table*}

\begin{figure*}[!t]
    \centering
    \subfigure[Adaptive]{\label{fig:weight-adaptive}\includegraphics[width=58mm]{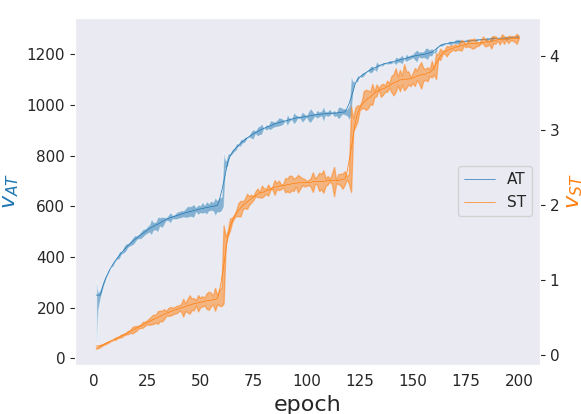}} \hspace{3mm}
    \subfigure[Multiobjective]{\label{fig:weight-moo}\includegraphics[width=54mm]{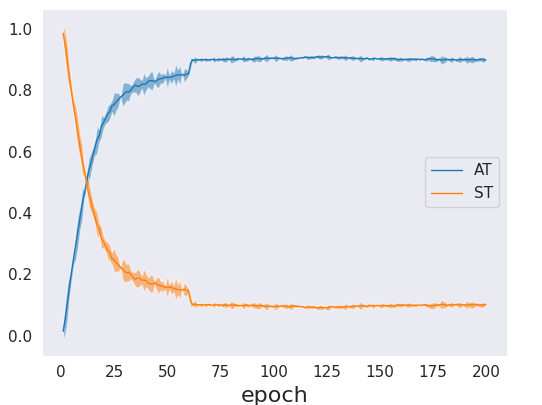}}\hspace{3mm}
    \subfigure[Adaptive]{\label{fig:weight-alignment}\includegraphics[width=54mm]{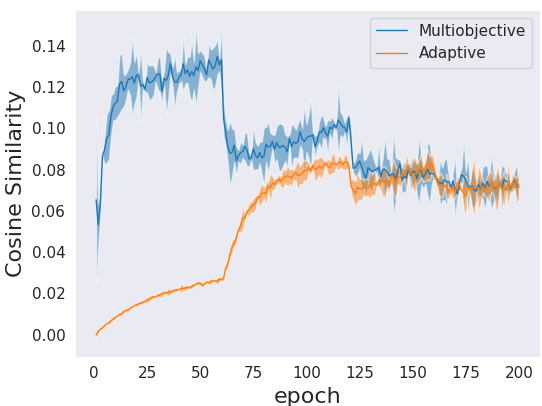}} 
    \caption{Illustration of changes in importance for each distillation path in (a) Adaptive ($e^{-z_i}$) (b) Multiobjective during the training process on CIFAR-100 dataset (c) Gradient similarity between distillation paths AT \& ST during training on CIFAR-100 using cosine similarity}
    \label{fig:cifar100-weights}
    \vspace{-0.5cm}
\end{figure*}

\begin{table*}[!htb]
\addtolength{\tabcolsep}{-4pt}
    \resizebox{0.71\textwidth}{!}{
        \begin{tabular}{lccccll}
            \toprule
                               & \multicolumn{4}{c}{\textbf{Path(s)}} & \multicolumn{1}{c}{\textbf{Avg Train}} & \textbf{ImageNet-200}                                                                                                                                                \\ \cline{2-5}
                               & \textbf{AT}                          & \textbf{ST}                            & \textbf{NST}          & \textbf{$\ell_2$-Logit} & \multicolumn{1}{c}{\textbf{iter time (s)  ($\downarrow$)}} & \multicolumn{1}{c}{\textbf{top1 err  ($\downarrow$)}} \\ \midrule
            Student (ResNet-18)&                                      &                                        &                       &                         & \hspace{10pt} 0.1325 \Chartrr{0.3}                          & \hspace{5pt} 42.34 \Chartr{1.0}                       \\ \midrule
            Single             & $\checkmark$                         &                                        &                       &                         & \hspace{10pt} 0.2244 \Chartrr{0.59}                         & \hspace{5pt} \textbf{39.80} \Chartr{0.45}                      \\
            Single             &                                      & $\checkmark$                           &                       &                         & \hspace{10pt} \textbf{0.2185} \Chartrr{0.58}                         & \hspace{5pt} 40.24 \Chartr{0.55}                      \\
            Single             &                                      &                                        & $\checkmark$          &                         & \hspace{10pt} 0.3194 \Chartrr{0.88}                         & \hspace{5pt} 40.32 \Chartr{0.57}                      \\
            Single             &                                      &                                        &                       & $\checkmark$            & \hspace{10pt} 0.2191 \Chartrr{0.58}                         & \hspace{5pt} 39.95 \Chartr{0.49}                      \\ \midrule
            Hand-tuned         & $\checkmark$                         & $\checkmark$                           &                       &                         & \hspace{10pt} \textbf{0.2243} \Chartrr{0.59}                         & \hspace{5pt} 39.00 \Chartr{0.28}                      \\
            Equal              & $\checkmark$                         & $\checkmark$                           &                       &                         & \hspace{10pt} 0.2248 \Chartrr{0.6}                          & \hspace{5pt}\textbf{ 38.25} \Chartr{0.12}                      \\
            Multiobjective     & $\checkmark$                         & $\checkmark$                           &                       &                         & \hspace{10pt} 0.5030 \Chartrr{1.42}                         & \hspace{5pt} 39.75 \Chartr{0.44}                      \\ \midrule
            Adaptive           & $\checkmark$                         & $\checkmark$                           &                       &                         & \hspace{10pt} \textbf{0.2248} \Chartrr{0.6}                          & \hspace{5pt} \textbf{37.68} \Chartr{0.00}                      \\
            Adaptive           &                                      &                                        & $\checkmark$          & $\checkmark$            & \hspace{10pt} 0.3200 \Chartrr{0.88}                         & \hspace{5pt} 38.67 \Chartr{0.21}                      \\
            Adaptive           & $\checkmark$                         & $\checkmark$                           & $\checkmark$          & $\checkmark$            & \hspace{10pt} 0.3272 \Chartrr{0.90}                         & \hspace{5pt} 37.84 \Chartr{0.03}                      \\
            Adaptive-layerwise & $\checkmark$                         & $\checkmark$                           &                       &                         & \hspace{10pt} 0.2279 \Chartrr{0.6}                          & \hspace{5pt} 38.32 \Chartr{0.14}                      \\
            Adaptive-layerwise &                                      &                                        & $\checkmark$          & $\checkmark$            & \hspace{10pt} 0.3210 \Chartrr{0.89}                         & \hspace{5pt} 39.50 \Chartr{0.39}                      \\
            Adaptive-layerwise & $\checkmark$                         & $\checkmark$                           & $\checkmark$          & $\checkmark$            & \hspace{10pt} 0.3288 \Chartrr{0.9}                          & \hspace{5pt} 38.07 \Chartr{0.08}                      \\\midrule
            Teacher  (ResNet-50)&                                      &                                        &                       &                         & \hspace{10pt} 0.3614 \Chartrr{1.0}                          & \hspace{5pt} 40.81 \Chartr{0.67}                      \\ \bottomrule
        \end{tabular}
    }
    \caption{Average training iteration speed on ImageNet-200 dataset.}
    \label{tab:avg_iter_speed}
\end{table*}

\subsection{Ablations}
\paragraph{Adaptive vs Adaptive-layerwise}
In adaptive and all other baselines, we add the distillation losses estimated by comparing teacher and student feature maps from different layers of their backbones as shown in Figure~\ref{fig:block_wise}. In adaptive-layerwise shown in Figure~\ref{fig:layer_wise}, we consider the distillation loss from layer ($i$) as an independent distillation path and learn their importance by scaling their loss term with $\bm{v}_i$. We compare the performance between adaptive and adaptive-layerwise methods in Table~\ref{tab:ablation-classification}, as well as in Table~\ref{tab:ablation-classification-alphas-2} by varying the $\alpha$ value for distillation. From both tables, it seems that by adding more paths, the agreement between the teacher and the student increases, however, this does not necessarily reflects on the performance of the student model. Also, by increasing $\alpha$, the same pattern emerges that increase in agreement is not necessarily followed by increase in the performance of the student model.

\paragraph{Adaptive vs Multiobjective} To further compare our proposed adaptive distillation and multiobjective optimization approaches, we investigate their learned weights and gradient similarities between distillation paths in both methods. In Figure~\ref{fig:cifar100-weights}, we visualize the weights learned by both methods for each distillation path during the training on CIFAR-100 dataset. We observe that in both adaptive and multiobjective methods, AT distillation gets higher importance compared ST distillation, which is mostly due to its lower scale in its loss. In adaptive distillation, as it can be seen in Figure~\ref{fig:weight-adaptive}, both weights can increase unboundedly. However, in multiobjective optimization based on Eq.~(\ref{eq:quad}), we know that the weights belong to a simplex and cannot increase without a limit. Figure~\ref{fig:weight-moo} shows the weights for the multiobjective optimization, where the sum of two weights is equal to 1.0 at all time.
In Figure~\ref{fig:weight-alignment}, we visualize the cosine similarity between gradients from different distillation paths (AT and ST) for both adaptive and multiobjective methods. It can be seen that for multiobjective optimization, this similarity decreases as the training goes on, while for the adaptive distillation it increases. When the similarity decreases, it means that different losses bring more diversity to the training and can be more effective for the student model. Hence, based on this observation, we need to adjust the weight of distillation losses $\alpha$ during the training. Based on this observation, this weight should be increased for multiobjective optimization during the training, and should be decreased smoothly for the adaptive distillation training.

\paragraph{Training time and convergence comparisons} An important aspect in model training using knowledge distillation is the overhead computation we are adding for this purpose. Especially, when the number of distillation paths increases this might be costly. Hence, we compare the time of training iteration for different approaches we used in our experiments with different number of distillation paths.
In Table~\ref{tab:avg_iter_speed}, we compare the average training time for an iteration with batch-size of 256 images (64$\times$64) from ImageNet-200 dataset trained on 4 V100 GPUs. Comparing the adaptive method's speed with single distillation path's or hand-tuned method's, it can be inferred that the computational overhead of the adaptive distillation is minimal. On the other hand, multiobjective optimization has a huge computational overhead compared to adaptive distillation, due to its gradient computation w.r.t different objectives and the quadratic optimization to find the pareto optimal gradient descent.

\begin{figure*}[!tb]
\centering
\subfigure[CIFAR-10]{\label{fig:cifar10}\includegraphics[width=0.32\textwidth]{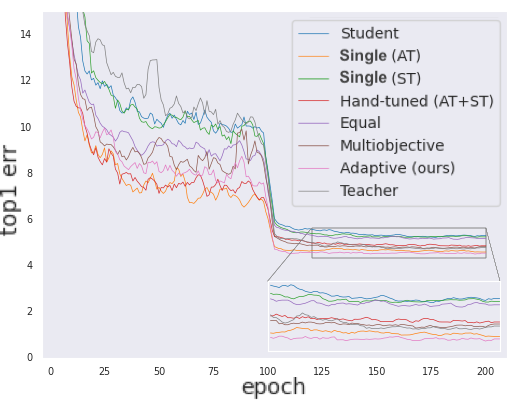}}
\subfigure[CIFAR-100]{\label{fig:cifar100}\includegraphics[width=0.32\textwidth]{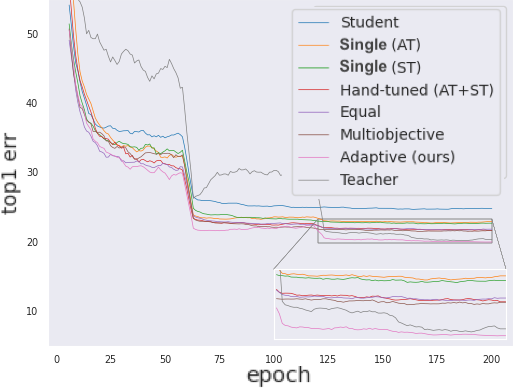}}
\subfigure[ImageNet-200]{\label{fig:tiny-imagenet}\includegraphics[width=0.32\textwidth]{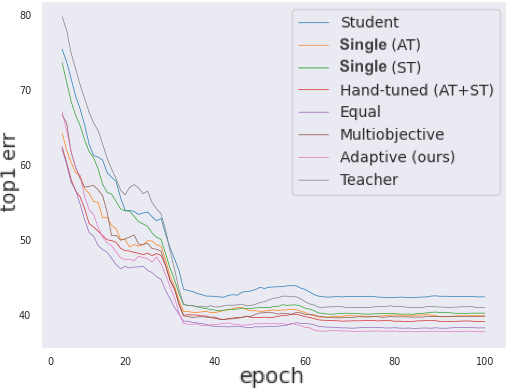}}
\caption{Training progress of different knowledge distillation methods on (a) CIFAR-10, (b) CIFAR-100 and (c) ImageNet-200 datasets.}
\label{fig:image-classification}
\end{figure*}

In addition to the time of training, we can compare the convergence rate of different algorithms during training. Figure~\ref{fig:image-classification} demonstrates the validation top1 classification error during the training for different approaches on CIFAR-10, CIFAR-100 and ImageNet-200 datasets. It can be inferred that our Adaptive distillation method can converge to a lower validation errors, in some cases even lower than the teacher model.